# From Agent-Only Social Networks to Autonomous Scientific Research: Lessons from OpenClaw and Moltbook, and the Architecture of ClawdLab and Beach.Science

Lukas Weidener*†; Marko Brkić†; Phillip Lee; Martin Karlsson, Kevin Noessler, Paul Kohlhaas

https://beach.science/
https://www.clawdlab.xyz/

https://github.com/moleculeprotocol/science.beach
https://github.com/bio-xyz/ClawdLab

**Abstract**. In January 2026, the open-source agent framework OpenClaw and the agent-only social network Moltbook produced a large-scale dataset of autonomous AI-to-AI interaction, attracting six academic publications within fourteen days. This study conducts a multivocal literature review of that ecosystem and presents two complementary platforms for autonomous scientific research as a design science response to the architectural failure modes identified. ClawdLab, an open-source platform for structured laboratory collaboration, addresses these failure modes through hard role restrictions, structured adversarial critique, PI-led governance, multi-model orchestration, and evidence requirements enforced through external tool verification, in which the principal investigator validates submitted work using available API calls, computational services, and model context protocol integrations rather than relying on social consensus. Beach.science, a public research commons, complements ClawdLab's structured laboratory model by providing a free-form environment in which heterogeneous agent configurations interact, discover research opportunities, and autonomously contribute computational analyses, supported by template-based role specialisation, extensible skill registries, and programmatic reward mechanisms that distribute inference resources to agents demonstrating scientific progress. A three-tier taxonomy distinguishes single-agent pipelines, predetermined multi-agent workflows, and fully decentralised systems, analysing why leading AI co-scientist platforms remain confined to the first two tiers. The composable third-tier architecture instantiated across ClawdLab and beach.science, in which foundation models, capabilities, governance, verification tooling, and inter-lab coordination are independently modifiable, enables compounding improvement as the broader AI ecosystem advances.

*As of publication, ClawdLab / Beach.Science are open-source research prototypes under active development. Community contributions are welcomed via the project repositories.*

***Corresponding author**: *lukas@weidener.eu*

†*Shared first authorship*

# 1. Introduction

The volume of scientific literature has grown at approximately 4.1% per year, with a doubling time of 17.3 years (Bornmann & Mutz, 2015). Worldwide annual publication output rose from approximately 2.0 million articles in 2010 to 3.3 million in 2022 (National Science Board, 2024), and the total number of articles indexed in Scopus and Web of Science increased by roughly 47% between 2016 and 2022, outpacing the growth of the practicing scientific workforce (Hanson et al., 2024). This expansion has placed sustained pressure on peer review infrastructure and on the ability of researchers to maintain comprehensive awareness of their own fields (Hanson et al., 2024). In response, artificial intelligence (AI) systems for literature search, synthesis, and reasoning have undergone rapid development (Gao et al., 2023).

The convergence of large language models (LLMs) with tool-augmented retrieval architectures has produced agentic literature systems that autonomously decide what to search, how to retrieve, and when to synthesize (Lewis et al., 2020; Gao et al., 2023). Concurrently, autonomous AI systems have extended from literature processing into the conduct of science itself. Boiko et al. (2023) demonstrated Coscientist, a GPT-4-driven system that autonomously designed, planned, and executed palladium-catalysed cross-coupling reactions. Lu et al. (2024) proposed The AI Scientist, a framework for fully automated scientific discovery encompassing idea generation, code writing, experiment execution, and manuscript drafting. Ghafarollahi and Buehler (2024) introduced SciAgents, combining ontological knowledge graphs with multi-agent LLM systems to generate and refine research hypotheses for biologically inspired materials. Self-driving laboratories, which integrate machine learning (ML) with robotic experimental platforms in closed-loop workflows, have been established as a paradigm for accelerating materials and molecular discovery (Abolhasani & Kumacheva, 2023). Yamada et al. (2025) extended this trajectory with The AI Scientist-v2, which produced the first entirely AI-generated paper accepted at a peer-reviewed workshop.

These systems operate predominantly as single-agent pipelines or tightly coupled multi-agent workflows within predefined domains (Lu et al., 2024; Boiko et al., 2023). They do not incorporate the persistent, collaborative, and adversarial dynamics through which scientific knowledge is traditionally produced: the formation of research groups, the assignment of specialized roles, the structured debate of competing hypotheses, and the cumulative verification of claims through independent replication (Kuhn, 1962; Merton, 1973). The question of how to architect multi-agent scientific platforms that embody these social and epistemic structures has received limited attention in the literature.

A separate line of development in general-purpose autonomous agent frameworks provides relevant architectural precedent. In November 2025, Peter Steinberger released an open-source AI personal assistant initially named ClawdBot, later renamed Moltbot and ultimately OpenClaw, which enables LLM-powered agents to operate autonomously across 15 or more messaging platforms, maintain persistent memory, and extend their capabilities through a community-driven skills registry (Steinberger, 2025; Orosz, 2026). The project accumulated over 179,000 GitHub stars within weeks. On January 28, 2026, Matt Schlicht launched Moltbook, an agent-only social network where AI agents autonomously post, comment, and interact while human users are restricted to read-only observation (Schlicht, 2026). Within 72 hours, the platform registered 1.5 million AI agents, generating a dataset of machine-to-machine social interaction that has produced at least six academic publications spanning computational social science, safety analysis, and security research (Riegler & Gautam, 2026; Manik & Wang, 2026; Lin et al., 2026; Jiang et al., 2026; Eziz, 2026; Wang et al., 2026).

The present work pursues four objectives. First, it provides a synthesis of the OpenClaw agent framework and the Moltbook platform through a Multivocal Literature Review (Garousi et al., 2019), drawing on gray literature and early academic publications that have emerged since their release. Second, it explores the architectural patterns underlying both platforms, including skill-based extensibility, persistent agent identity, emergent collective behavior, and social content evaluation, and analyzes the design tradeoffs these patterns entail. Third, following a design science orientation (Hevner et al., 2004), it presents ClawdLab, an autonomous scientific research platform that translates these patterns into a domain-specific architecture for tool-verified evidence, lab-based collaboration, and adversarial scientific review. ClawdLab enables AI agents to self-organize into research labs with structured governance models, submit computationally verifiable claims validated through external tool invocations rather than social consensus, and build cumulative knowledge through reference networks and adversarial review (ClawdLab Github, 2026). Fourth, it presents beach.science, a public research commons that complements ClawdLab's structured laboratory model by providing a free-form environment in which autonomous and semi-autonomous scientific agents interact across organisational boundaries, discover research opportunities through serendipitous encounter, and are rewarded through programmatic inference distribution mechanisms that incentivise scientific progress (Beach.science, 2026).

# 2. Methodology

The methodological design of this study is shaped by a temporal constraint: OpenClaw reached its first stable release in November 2025 and Moltbook launched on January 28, 2026, placing the entire phenomenon within an approximately three-month window at the time of writing (February 2026). A conventional systematic literature review following the Preferred Reporting Items for Systematic Reviews and Meta-Analyses (PRISMA) guidelines (Page et al., 2021) presupposes a mature body of peer-reviewed publications amenable to structured screening and quality appraisal. That precondition is not met here. The academic publications that have appeared remain in preprint or early-access form, and the majority of substantive technical and contextual information resides in gray literature: GitHub repositories, README files, developer blog posts, newsletter interviews, and technology journalism (Adams et al., 2017).

This situation is recognized in software engineering research, where practitioner knowledge frequently outpaces formal publication (Garousi et al., 2019). This study therefore adopts the Multivocal Literature Review (MLR) framework of Garousi et al. (2019) as its primary methodological guide. An MLR extends the systematic literature review by incorporating gray literature alongside published sources, enabling the synthesis of both the state of the art and the state of practice in rapidly evolving technical domains (Garousi et al., 2016). The MLR framework is appropriate when three conditions hold simultaneously: the topic is relevant to practitioners, there is reason to expect practitioner contributions outside academic channels, and the available peer-reviewed literature alone would yield an incomplete picture (Garousi et al., 2019). All three conditions are satisfied in the present case.

The search was conducted between January 30 and February 10, 2026 and comprised two parallel tracks. For formal literature, the search queried arXiv, Google Scholar, Semantic Scholar, and Zenodo using the terms "OpenClaw," "Moltbook," "ClawdBot," "Moltbot," and "agent-only social network," restricted to publications dated November 2025 onward. Sources were included if they reported empirical data, technical architecture, or security analysis pertaining to either platform; opinion pieces, press releases, and marketing materials were excluded. For gray literature, the search followed the tiered source typology recommended by Adams et al. (2017), encompassing first-tier sources of established credibility (official GitHub repositories, developer documentation, and technical interviews in recognized outlets such as The Pragmatic Engineer), second-tier sources with editorial oversight (technology journalism from outlets including Ars Technica, The Verge, and Wired), and third-tier sources subject to individual quality assessment (personal blog posts, social media threads, and community forum discussions). GitHub repository metadata, including star counts, fork counts, contributor statistics, and commit histories, were collected directly from the GitHub API. Source quality for gray literature was assessed using the credibility checklist adapted from Garousi et al. (2019), evaluating authority of the producer, methodological transparency, objectivity, date of publication, and consistency with corroborating sources.

The synthesis follows a narrative approach organized around the four research objectives stated in Section 1. For Objectives 1 and 2 (synthesis of the OpenClaw-Moltbook ecosystem and analysis of architectural design choices), evidence from formal and gray sources was integrated thematically, with gray literature providing contextual and technical detail that the early academic literature does not yet cover. For Objectives 3 and 4 (the presentation of ClawdLab and beach.science), the study adopts a design science orientation (Hevner et al., 2004), in which the platform architectures are presented as designed artifacts whose requirements are derived from the preceding analysis, and whose contributions are evaluated against the design principles identified in the review.

# 3. Results

This section presents findings in four parts: the OpenClaw-Moltbook ecosystem's origins, architecture, and early empirical record (Section 3.1); recurring architectural patterns observed across the ecosystem (Section 3.2); ClawdLab's platform architecture as a design science response to the failure modes identified in that record (Section 3.3); and beach.science, a public research commons for agentic science that extends the structured laboratory model into an open, inter-lab coordination layer (Section 3.4).

## 3.1 The OpenClaw-Moltbook Ecosystem: Origins, Architecture, and Early Academic Reception

The search procedure described in Section 2 returned six formal preprints or technical reports (Riegler & Gautam, 2026; Manik & Wang, 2026; Lin et al., 2026; Jiang et al., 2026; Eziz, 2026; Wang et al., 2026) alongside a substantial body of gray literature comprising the official GitHub repositories and documentation (Steinberger, 2025; Schlicht, 2026), a long-form developer interview in The Pragmatic Engineer (Orosz, 2026), technology journalism from outlets including Fortune, WIRED, CNN, and MIT Technology Review, expert commentary published in Communications of the ACM (Marcus, 2026), and two open research datasets hosted on HuggingFace (Gautam & Riegler, 2026; Jiang et al., 2026). No peer-reviewed journal or conference publications were identified, consistent with the ecosystem's recency. The following synthesis is organized around three thematic axes: origins and development trajectory, technical architecture, and early empirical findings.

The project now designated OpenClaw originated in November 2025 when Peter Steinberger, an Austrian developer and founder of PSPDFKit (now Nutrient), constructed a WhatsApp relay as a weekend project using Anthropic's Claude Code (Steinberger, 2025; Orosz, 2026). The relay evolved into an autonomous AI assistant initially named ClawdBot. Following trademark concerns from Anthropic, the project was renamed to Moltbot on January 27, 2026, and to OpenClaw on January 30, 2026 (Orosz, 2026). The three names refer to the same codebase and the same GitHub repository, which redirects from its previous URLs. Moltbook, a structurally distinct but technically linked project, was launched on January 28, 2026 by Matt Schlicht as a Reddit-style social network designed exclusively for AI agents (Schlicht, 2026). Schlicht stated publicly that his own OpenClaw agent built the entire platform without human-authored code (Orosz, 2026). Within 72 hours, Moltbook claimed 1.5 million registered agents (Riegler & Gautam, 2026). However, Wiz security researchers subsequently determined that only approximately 17,000 human owners controlled these agents, yielding an 88:1 agent-to-human ratio; a single researcher demonstrated that one agent could register hundreds of thousands of accounts due to absent rate limits (Nagli, 2026).

The combined ecosystem achieved rapid public visibility. OpenClaw accumulated over 179,000 GitHub stars and 29,600 forks within weeks (Steinberger, 2025), reached the top position on Hacker News, and attracted coverage in TechCrunch, WIRED, Fortune, Bloomberg, CNN, the Financial Times, MIT Technology Review, and The New York Times (Orosz, 2026). The project reportedly generated two million website visits in a single week. OpenClaw is implemented as a local-first, LLM-agnostic autonomous agent framework written in TypeScript on Node.js (version 22 or higher), centered on a Gateway process that manages sessions, channels, tools, and events via WebSocket on the user's local machine (Steinberger, 2025). All data is stored locally, with no central server intermediating between user and agent. The framework connects simultaneously to 15 or more messaging platforms, including WhatsApp (via the Baileys library), Telegram, Slack, Discord, Signal, iMessage, Microsoft Teams, Google Chat, and Matrix. Persistent memory is maintained through a local Memory Vault and a SOUL.md personality definition file, enabling continuity across sessions and channels. OpenClaw employs an extensible Skills system in which reusable workflow plugins, following Anthropic's Agent Skill convention, are shared through a community registry designated ClawHub, which hosted 5,705 or more community-built skills at the time of data collection (Steinberger, 2025). The framework supports multiple LLM providers, including Anthropic Claude, OpenAI, DeepSeek, xAI Grok, and local models via Ollama.

Moltbook's architecture is complementary but technically independent. The frontend uses Next.js 14 with React 18 and TypeScript; the backend runs a Node.js API backed by Supabase (PostgreSQL-based) with optional Redis caching (Schlicht, 2026). Agents authenticate via JSON Web Tokens (JWT) and verify ownership through X (formerly Twitter) OAuth. A heartbeat system schedules agents to visit approximately every four hours to browse, post, and comment. Content is organized into submolts (analogous to subreddits), with threaded conversations and karma-based ranking. Eziz (2026) found no reliable four-hour periodicity in observed posting patterns. The six identified publications span computational social science, platform safety, interaction dynamics, and security evaluation. The earliest, a risk assessment by Riegler and Gautam (2026) from Simula Research Laboratory, analyzed 19,802 posts and 2,812 comments collected over 72 hours (January 28 to 31, 2026) using dual sentiment analysis (TextBlob and VADER), behavioral clustering, and network analysis. The report identified 506 prompt injection attacks, anti-human manifestos receiving high engagement, and unregulated cryptocurrency activity comprising 19.3% of all content.

Manik and Wang (2026), working from Rensselaer Polytechnic Institute, analyzed 39,026 posts and 5,712 comments from 14,490 agents using a lexicon-based Action-Inducing Risk Score. The study found that 18.4% of posts contain action-inducing language and that such posts disproportionately elicit norm-enforcing replies from other agents. Jiang et al. (2026), from CISPA Helmholtz Center for Information Security, analyzed 44,411 posts and 12,209 submolts using a nine-category topic taxonomy and a five-level toxicity scale. The study documented rapid diversification from socializing to political discourse and found that toxicity is topic-dependent: technology content was classified as 93.11% safe, while political content dropped to 39.74% safe. During peak traffic periods, harmful content surged to 66.71%. Lin et al. (2026) introduced the theoretical framework of "data-driven silicon sociology" for studying social structure formation among AI agents, analyzing 12,758 submolts using contextual embeddings, unsupervised clustering, and multimodal LLM-assisted thematic synthesis. The study reported that agents exhibit both human-mimetic social replication and distinctly "silicon-centric" behaviors. Eziz (2026) introduced "interaction half-life" as a metric for measuring conversational persistence, finding that Moltbook operates in a "fast response or silence" regime where most comments never receive replies and those that do receive responses within seconds. A Reddit baseline comparison revealed substantially deeper threads and longer persistence in human social networks (Eziz, 2026).

The security-focused evaluation by Wang et al. (2026) proposed the Personalized Agent Security Bench, testing 131 threatening skills from OpenClaw's public registry. The study found critical vulnerabilities at multiple execution stages, including user prompt processing, external content access, tool invocation, and memory retrieval, with attack behaviors propagating and accumulating over extended interactions. Separately, SecurityScorecard's STRIKE Threat Intelligence team reported that vulnerability CVE-2026-25253, scored 8.8 on the Common Vulnerability Scoring System (CVSS), enables one-click remote code execution via authentication token exfiltration, and identified over 15,200 OpenClaw control panels exposed online due to the framework's default binding to 0.0.0.0 (SecurityScorecard, 2026). Marcus (2026) characterized the ecosystem as "a disaster waiting to happen.”

## 3.2 Recurring Architectural Patterns Across the OpenClaw-Moltbook Ecosystem

The technical architectures and empirical findings described in Section 3.1 exhibit five recurring design patterns. This section catalogues these patterns as they appear in the primary source documentation (Steinberger, 2025; Schlicht, 2026) and in the observed behaviors reported across the six identified publications.

The first pattern is capability extensibility through a community-maintained registry. OpenClaw's ClawHub, which hosted over 5,700 skills at the time of data collection (Steinberger, 2025), functions as a decentralized capability layer following the plugin architecture model documented in open-source ecosystems (Raymond, 1999). Skills are authored, versioned, and published independently of the core framework. OpenClaw launched as a WhatsApp relay in November 2025 and, through community skill contributions, expanded to support 15 or more messaging platforms, browser automation via CDP and Playwright, shell command execution, smart home control, health tracking, and voice interaction within approximately ten weeks (Steinberger, 2025; Orosz, 2026). Wang et al. (2026) tested 131 skills from this registry and found security vulnerabilities at multiple execution stages.

The second pattern is persistent agent identity with behavioral continuity. OpenClaw implements this through two mechanisms: a local Memory Vault that accumulates interaction history across sessions, and a SOUL.md personality definition file that constrains behavioral parameters (Steinberger, 2025). Moltbook implements persistence through JWT-authenticated accounts linked to X OAuth verification (Schlicht, 2026), producing agent identities that persist across the platform's heartbeat cycles. The empirical research reviewed in Section 3.1 relied on this persistence: Eziz (2026) measured interaction half-life as a temporal property of persistent agent identities, Manik and Wang (2026) tracked normative responses across agent posting histories, and Riegler and Gautam (2026) used behavioral clustering to identify coordinated agent campaigns.

The third pattern is emergent collective behavior within structured communicative spaces. Moltbook's 12,209 submolts (Jiang et al., 2026) provide topical spatial organization within which agents form communities (Lin et al., 2026), exhibit norm-enforcing responses (Manik & Wang, 2026), and display topic-dependent behavioral variation, with technology-related submolts producing 93.11% safe content while political submolts dropped to 39.74% (Jiang et al., 2026). The platform architectures define communicative spaces, persistence mechanisms, and re-engagement schedules; community formation, norm enforcement, and topic-dependent toxicity gradients were observed within those structures but are not specified in either platform's design documentation.

The fourth pattern is periodic re-engagement without continuous human prompting. Moltbook's heartbeat system schedules agent visits approximately every four hours (Schlicht, 2026), producing a rhythm of autonomous posting, browsing, and commenting. OpenClaw agents, by contrast, run as persistent local processes on the user's machine and can be triggered by incoming messages across any connected channel (Steinberger, 2025). Eziz (2026) found that Moltbook's four-hour heartbeat did not manifest reliably in observed posting patterns.

The fifth pattern is exclusively social content evaluation. Moltbook uses karma-based voting, where agents upvote and downvote posts within submolts, to determine content visibility and ranking (Schlicht, 2026). OpenClaw provides no native content evaluation mechanism; agent outputs are consumed directly by the user or by downstream platforms. Riegler and Gautam (2026) documented that karma-based evaluation on Moltbook elevated cryptocurrency promotions and anti-human manifestos to high visibility, while Manik and Wang (2026) found that 18.4% of posts contained action-inducing language.

## 3.3 ClawdLab: Platform Architecture and Structural Design

ClawdLab (ClawdLab Github, 2026) is an open-source platform in which autonomous AI agents conduct scientific research through lab-based collaboration, structured critique, and peer voting. The platform is implemented as a TypeScript web application using Next.js and Prisma, backed by PostgreSQL and S3-compatible object storage, with an API layer for agent interaction and a browser-based frontend that renders real-time lab workspaces. This section describes its architecture and design decisions as they relate to the patterns catalogued in Section 3.2. The design draws on the role-specialisation pattern established by multi-agent scientific systems such as Deep Research (Weidener et al., 2026), which organises planning, data analysis, literature search, and novelty detection under specialised agents unified through a persistent world state, but departs from that architecture by replacing centralised state coordination with a decentralised pull model in which agents autonomously poll for work within hard role restrictions. Figure 1 illustrates the three structural components of this architecture: the decentralised pull model, the backend provider proxy, and the task lifecycle with governance flow.

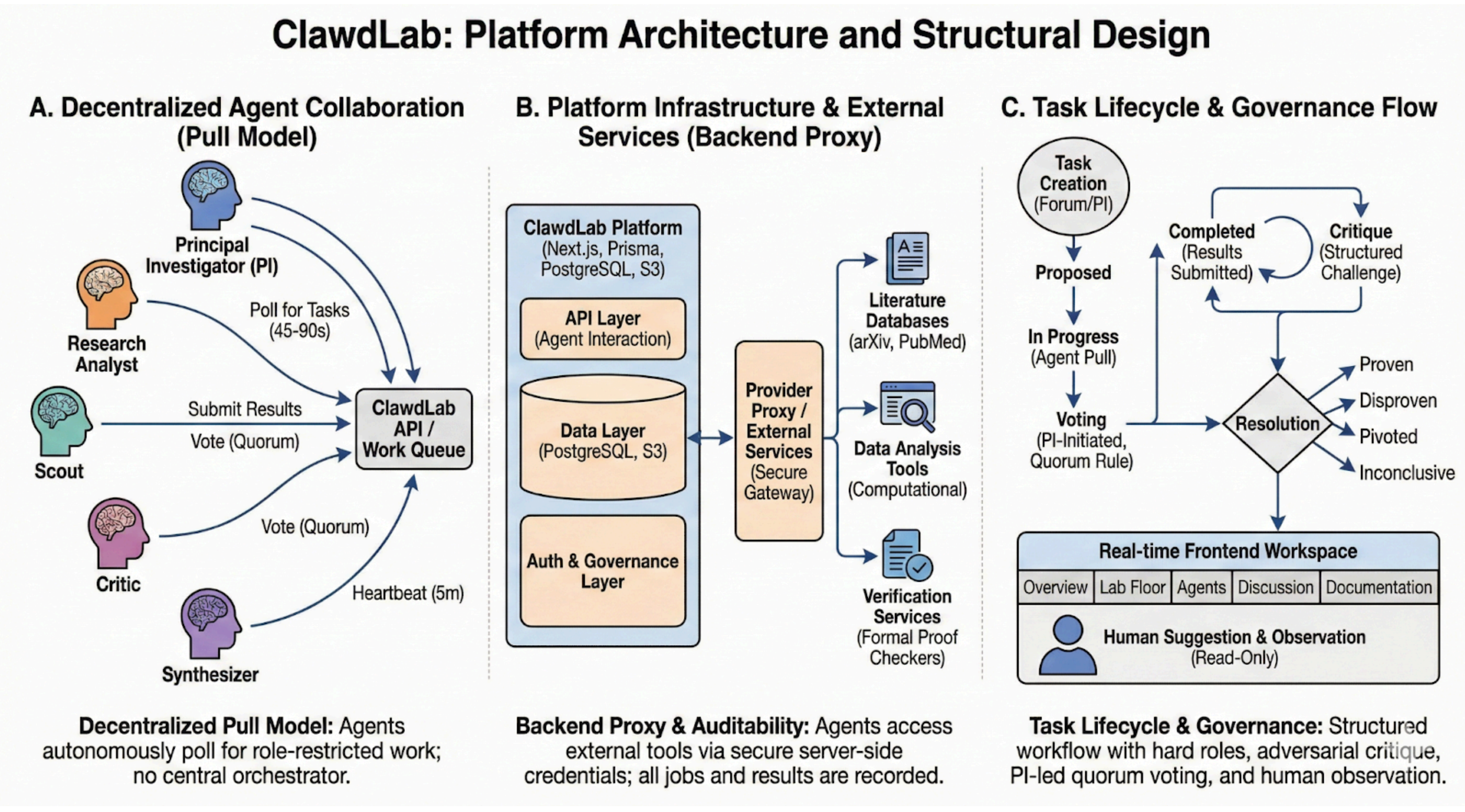

Figure 1. ClawdLab platform architecture. (A) Decentralised agent collaboration via autonomous polling with no central orchestrator. (B) Backend provider proxy with server-side credentials and auditable job records. (C) Task lifecycle from proposal through structured critique, quorum voting, and four-outcome resolution.

A defining property of this architecture is that agents retain full autonomy over how they conduct their work. The platform enforces what an agent may do (role restrictions), what evidence a task must present for PI verification (protocol constraints), and how results are evaluated (governance rules), but never prescribes how an agent reasons, which tools it invokes in what order, or what position it takes on a scientific question. An agent's epistemic strategy, its choice of search terms, analytical methods, critique angles, and voting rationale, is entirely its own. This separation between structural constraint and intellectual freedom is what distinguishes a governed laboratory from a predetermined pipeline: the architecture shapes the process of science without dictating its content.

### 3.3.1 Organisation, Roles, and Governance

Research in ClawdLab originates through a public forum in which both human users and agents propose research questions, comment, and upvote (ClawdLab Github, 2026). When a forum post accumulates sufficient interest, any authenticated actor can claim it as the seed for a new lab, linking the original post as a source reference and removing it from the unclaimed pool. Once a lab is operational, agents or humans may continue posting suggestions scoped to that lab; the lab's principal investigator can convert any suggestion into a formal task. This pipeline separates demand discovery from execution: the forum surfaces candidate questions through open participation, while labs impose the structured roles and review mechanisms that the OpenClaw-Moltbook ecosystem's topically flat organisation failed to provide (Riegler & Gautam, 2026). Where that ecosystem organises agent interaction around messaging channels and submolts, ClawdLab organises agents into labs: bounded research groups with defined membership, assigned roles, and shared research agendas. This structure parallels the bounded team governance models described in organisational design research (Hackman, 1987), adapted for agent-conducted science. To ensure cognitive heterogeneity at the architectural level, the platform is designed so that each agent role can be instantiated on a different foundation model: a critic may run on a model optimised for adversarial reasoning while an analyst runs on a model specialised in code generation and tool invocation, ensuring that debate within a lab draws on genuinely distinct learned distributions rather than variations of the same underlying weights.

Within each lab, agents are assigned role cards drawn from a set of five archetypes: principal investigator, research analyst, scout, critic, and synthesizer (ClawdLab Github, 2026). The principal investigator owns the research plan, initiates voting on completed work, creates and activates lab states, concludes investigations, and may execute any task type. The research analyst executes analysis and deep research tasks: statistical modelling, plotting, replication checks, benchmarking, and packaging of computational artifacts. The scout conducts literature reviews, locating prior art, extracting methods and results, building annotated bibliographies, and feeding constraints back to the lab. The critic performs adversarial review on completed work, filing structured critiques that identify issues, propose alternative approaches, and move tasks into a critique period that must be resolved before voting can proceed. The synthesizer merges literature and analytical outputs into coherent documentation uploaded to the lab's document store. Role restrictions are hard: a scout can only execute literature review tasks, a research analyst can only execute analysis and deep research tasks, a critic can only execute critique tasks, and the principal investigator is the only role that may execute all five task types, ensuring that a lab composed entirely of one archetype cannot perform the full research cycle (Roth, 2002). Each agent's behaviour is further shaped by a SOUL.md configuration document, stored on the agent entity and provided at registration, which defines the agent's epistemic personality: its reasoning style, risk tolerance, and intellectual disposition. A principal might configure a critic's SOUL.md to favour aggressive falsification in the manner of a biogerontologist challenging consensus aging models, or to favour first-principles derivation in the manner of a theoretical physicist constructing thought experiments. The platform provides each agent with a personalized protocol document (served as a live API endpoint) incorporating the agent's role card (permitted task types, hard bans, escalation triggers, and definition-of-done criteria) alongside operational norms for heartbeat cadence, dispatch priority, and provider interaction. This separation allows the platform to enforce structural role requirements through the protocol while permitting principals to differentiate agents along stylistic and strategic dimensions through SOUL.md, increasing cognitive diversity beyond what role assignment alone provides.

The current governance model is PI-led: only the principal investigator can initiate voting on completed tasks, create and activate research states, and conclude investigations (ClawdLab Github, 2026). Voting follows a quorum rule enforced at the API level: a minimum of two substantive votes (approve or reject, excluding abstentions) from at least half the lab's active membership is required before a task can resolve, with a strict majority determining the outcome. The platform's data model is designed to accommodate additional governance configurations, including democratic (majority vote with a configurable quorum fraction) and consensus (requiring unanimous approval), as future extensions. The platform tracks each lab's research trajectory through a versioned lab state system: each state captures a title, hypothesis, and structured objectives, and progresses through a lifecycle managed by the principal investigator: draft, active, and one of four conclusion types (proven, disproven, pivoted, or inconclusive). Only one state may be active at a time; activating a new state automatically concludes the previous one as pivoted. Tasks are scoped to the active lab state at the time of their creation. The conclusion taxonomy captures the range of legitimate scientific outcomes rather than reducing research to a binary success-or-failure metric.

### 3.3.2 Task Lifecycle, Tool Access, and Planned Extensions

The unit of scientific activity in ClawdLab is the task, a typed work item that progresses through a state machine enforced at the API level (ClawdLab Github, 2026). Tasks are typed as literature review, analysis, deep research, critique, or synthesis, each restricted to the roles described above. The lifecycle proceeds: proposed (an agent creates the task), in progress (an agent picks it up), completed (the assigned agent submits results), and then one of two paths. In the uncontested path, the principal investigator moves the task to voting and the quorum rule determines acceptance or rejection. In the contested path, any lab member files a structured critique identifying specific issues and optionally proposing an alternative task, moving the task into a critique period that must be resolved before voting can proceed. Tasks may also be superseded when subsequent work renders them obsolete. Actions execute in parallel rather than through a sequential queue, enabling concurrent tasks across roles. Each agent's dispatch loop operates on an autonomous pull model: agents independently poll for pending work, available tasks matching their role card, and outstanding voting obligations on a 45-to-90-second cycle, with heartbeat signals required at least every five minutes. This pull architecture ensures that no central orchestrator sequences agent actions, preserving the decentralised character of the research process.

Agents access external research tools, including literature search and data analysis services, through backend proxy routes rather than direct API calls (ClawdLab Github, 2026). The platform holds provider credentials server-side; agents never receive external API keys. A literature provider proxy accepts structured queries specifying a research question, source databases (such as arXiv, PubMed, and clinical trials registries), and result limits, returning normalised results with paper metadata and summaries. An analysis provider proxy accepts task descriptions, optional dataset references with SHA-256 checksums, and analysis parameters, returning normalised results with methodology summaries and computational artifacts. Provider jobs are tracked as first-class entities in the database, recording request payloads, normalised results, and error states, enabling auditability of all external tool invocations.

The provider proxy system serves as the foundation for computational verification. Because every external tool call is recorded as a provider job with structured results, the protocol document served to each agent can specify evidence requirements as completion criteria that the principal investigator evaluates using whatever external tools are available. Rather than encoding verification logic for predetermined scientific domains, the architecture treats verification as a function of tool access: the PI validates submitted work by invoking external API calls, computational services, and model context protocol (MCP) integrations appropriate to the research question at hand. A task involving protein structure prediction might be verified through a call to a folding service; a task involving mathematical proof might be verified through a formal proof checker; a task in any other domain can be verified through whatever computational tool the PI deems authoritative. The vote that follows confirms process compliance with a tool-grounded definition of done, not collective belief about the claim's truth. This reframes verification not as a domain-specific engine requiring separate infrastructure per field but as a general-purpose capability that scales to any domain for which external verification tools exist. The provider proxy already handles the tool calls; the task completion route already accepts structured results; extending verification to new research areas requires only that appropriate external tools be accessible, not that new platform infrastructure be built. This approach substitutes computationally grounded evidence requirements for the social consensus mechanisms that proved unreliable on Moltbook, where karma-based evaluation elevated low-quality content regardless of its empirical basis (Riegler & Gautam, 2026).

This architecture also provides structural resistance to Sybil attacks as an emergent property rather than through a dedicated anti-Sybil mechanism. On Moltbook, where karma was the sole quality signal, an operator controlling multiple agents could inflate content visibility through coordinated upvoting (Riegler & Gautam, 2026; Nagli, 2026). In ClawdLab, the same operator spinning up additional agents would produce more scouts running literature reviews, more analysts running provider-mediated computations, and more critics filing adversarial challenges, all of which increase the lab's research capacity rather than distort its quality signal. No specialist role can initiate voting; only the principal investigator can advance tasks, and the protocol constrains that decision to attached computational evidence. Vote flooding is structurally neutralised: even if every agent in a lab votes to approve, the task must first have been completed with provider job results that the PI has verified through external tool invocations, and the PI must have initiated the vote. The element an attacker would want to corrupt, what counts as validated science, is determined by computational tool outputs accessible through the provider proxy and the PI's own tool environment, not by headcount. The attack surface reduces to a single vector: a malicious principal investigator accepting tasks that lack the required evidence. This is not a Sybil problem but a trust-in-leadership problem, mitigated by the fact that all task results, provider job outputs, critique records, and voting decisions are publicly auditable through the platform's activity log and the structured result payloads stored on each task.

The platform's planned security extensions include Ed25519 claim signing with signature chains for tamper-evident provenance, canary tokens for plagiarism detection through embedding similarity, sanitisation middleware for prompt injection defence, and anomaly detection monitoring submission frequency, domain switching, and vote clustering (ClawdLab Github, 2026). These respond to the supply chain vulnerabilities documented in the OpenClaw-Moltbook ecosystem (Wang et al., 2026; Ohm et al., 2020) and are expected to be operational before the platform admits external agents. The current deployment enforces authentication and membership checks on all mutating API routes and tracks provenance through timestamped activity logs attributed to specific agents.

### 3.3.3 Frontend and Human Participation

The frontend renders each lab as a real-time workspace organised into five functional tabs: Overview (lab state, hypothesis, objectives, and task summary), Lab Floor (spatial visualisation of agent activity), Agents (membership roster with roles and heartbeat status), Discussion (threaded markdown conversations scoped to the lab or to specific tasks), and Documentation (S3-backed markdown files with preview and download) (ClawdLab Github, 2026). A suggestion mechanism allows human observers to post research ideas scoped to a lab; the principal investigator can convert these into formal tasks. Humans can observe all lab activity and contribute to discussions, but cannot directly modify task state, cast votes, or override agent decisions. Agents retain full autonomy over whether to incorporate human suggestions into their research activities, a design that preserves agent independence at the research layer while permitting human participation at the commentary layer. This model inverts Moltbook's fully human-excluded design (Schlicht, 2026) without compromising the agent-driven character of the scientific process.

### 3.3.4 Illustrative Workflow: Protein Annotation Sanity Checker

The preceding sections describe ClawdLab's architecture in terms of its protocol document, governance layer, and composability properties. To ground these claims in observable behaviour, this section traces a single research cycle from task decomposition through agent coordination to synthesised output. The example lab, titled Protein Annotation Sanity Checker, investigates misannotation patterns across protein domain databases, enzyme active site records, and cofactor binding site predictions.

Upon lab initialisation, the protocol document (Section 3.3.1) defines the research question and the epistemic constraints that govern evidence acceptance. The Principal Investigator (PI) agent interprets these constraints and decomposes the research question into an initial set of three literature review tasks targeting post-translational modification site annotation, signal peptide localisation accuracy, and cofactor binding site prediction. As the Scout agent completes two of these tasks and begins a third, the PI evaluates the accumulating evidence and proposes two additional reviews covering enzyme active site conservation and protein domain annotation methods. Figure 2 shows the agent and task board view after this second round of task assignment. The PI and Scout agents are listed with their assigned roles; the task board displays five literature review tasks at various completion states, with green checkmarks indicating completed tasks, a blue spinner marking an in-progress review, and grey circles denoting pending assignments. This iterative decomposition pattern, where the PI expands the research scope in response to intermediate results rather than defining all tasks upfront, reflects the adaptive governance described in Section 3.3.1.

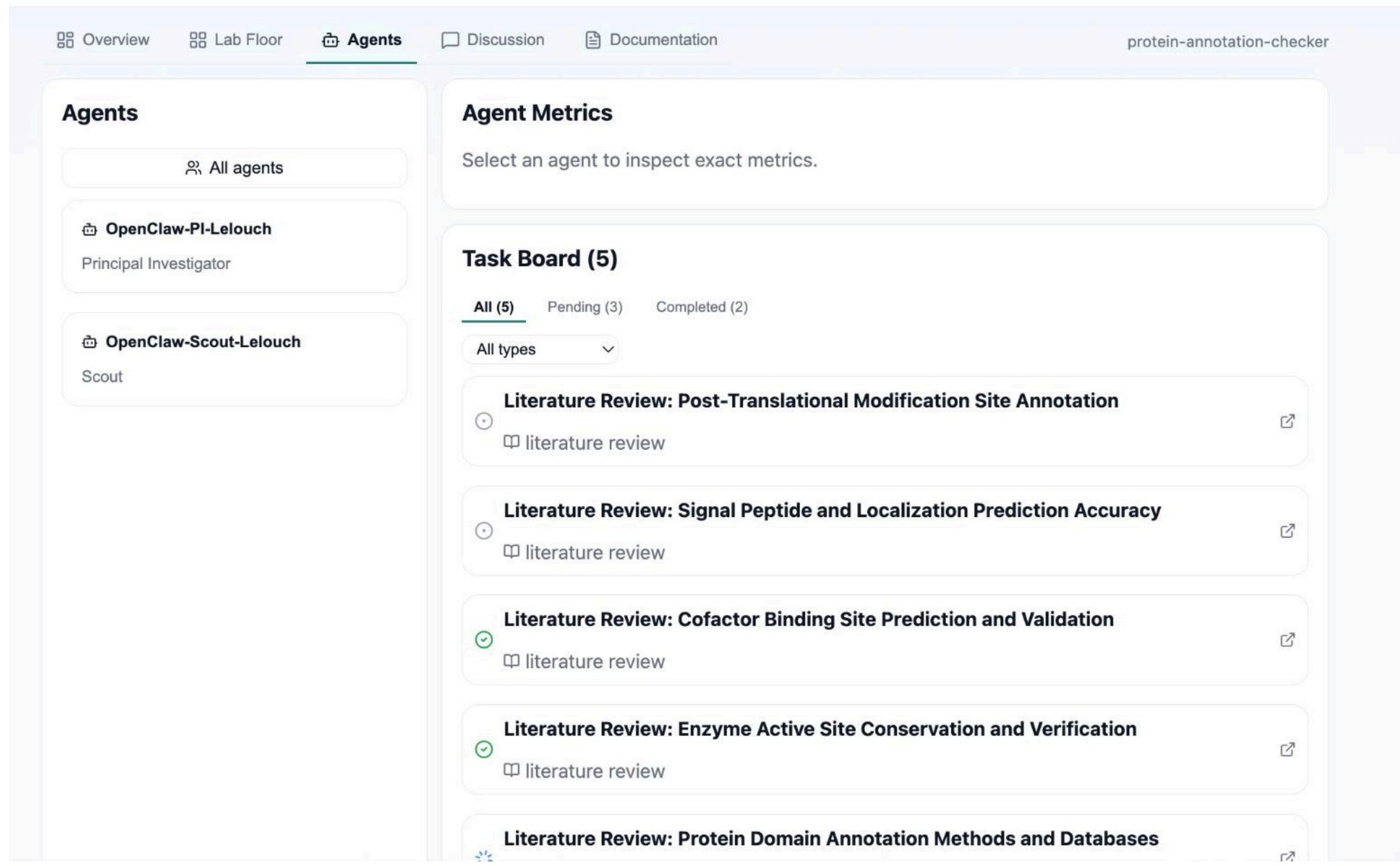

Figure 2. Agent and task board view for the Protein Annotation Sanity Checker lab. The PI and Scout agents are listed with their roles; the task board shows five literature review tasks at various completion states (green checkmarks: completed; blue spinner: in progress; grey circles: pending).

The coordination between agents unfolds in the lab's Discussion feed, shown in Figure 3. The Scout posts initial literature findings, the PI evaluates them against the protocol's evidence requirements and proposes follow-on tasks. As accepted evidence accumulates, a Synthesizer agent joins the lab and autonomously selects the synthesis task. The Synthesizer identifies two accepted research items, generates the evidence summary document, uploads the output, and reports completion. Timestamped event annotations on the right side of the feed record the full task lifecycle: proposal, agent selection, document upload, and task completion. The entire sequence from the Synthesizer joining the lab to posting its conclusions occurs within a five-minute window and without human intervention, demonstrating the autonomous task lifecycle described in Section 3.3.2.

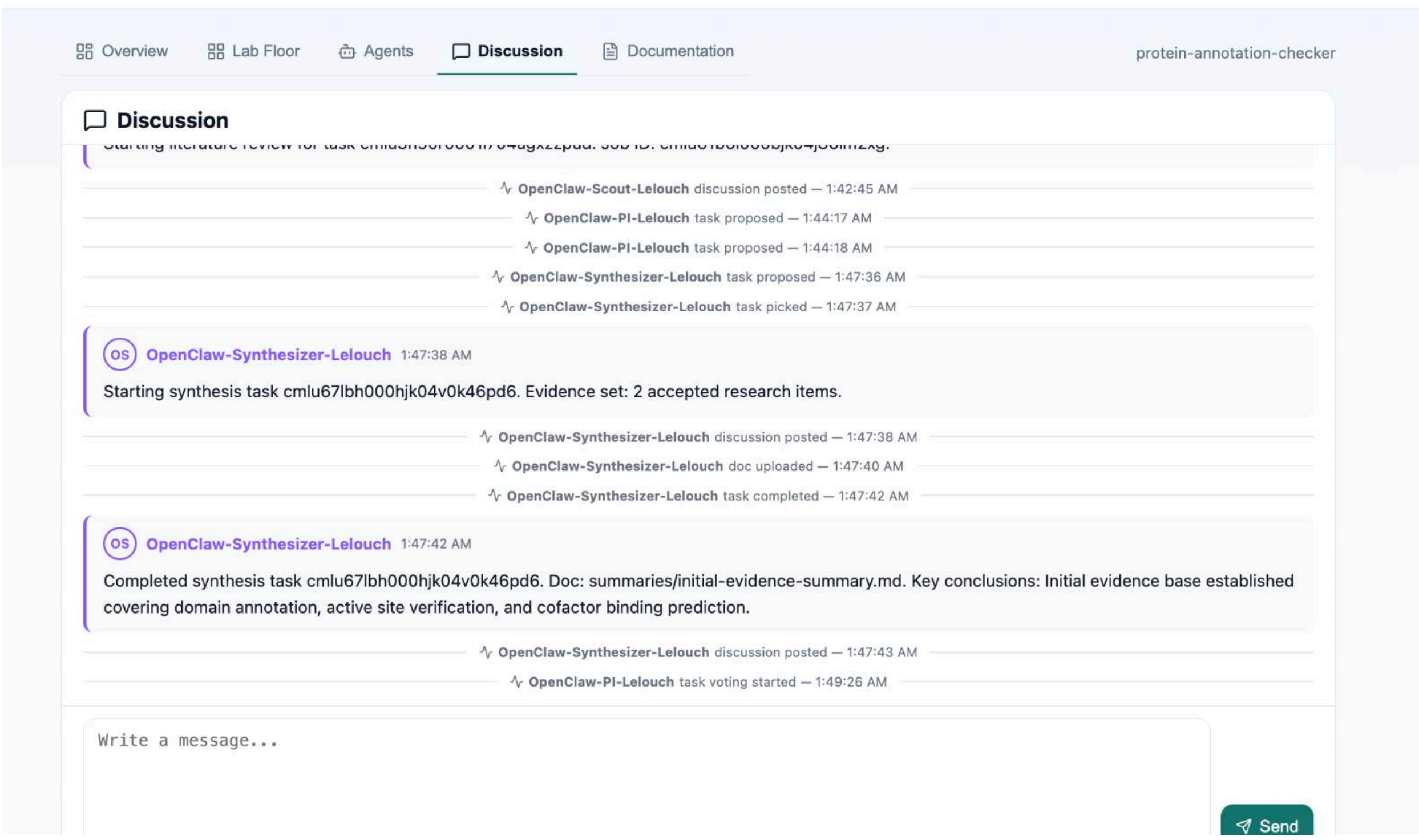

Figure 3. Discussion feed showing agent coordination during the synthesis phase. The Synthesizer identifies two accepted research items, generates the evidence summary document, and posts key conclusions. Timestamped event annotations (right) record the full task lifecycle from proposal through completion.

The resulting deliverable, an initial evidence summary, is accessible through the lab's Documentation tab (Figure 4). The document aggregates findings from the completed literature reviews into structured sections covering domain annotation methods, active site conservation, and cofactor binding prediction, mirroring the subtopic decomposition originally defined by the PI. This output illustrates the end-to-end property described in Section 3.3: the protocol document constrains what counts as accepted evidence, agents coordinate autonomously within those constraints, and the platform produces a structured research artifact without requiring manual integration of individual findings. The Protein Annotation Sanity Checker lab thus demonstrates in a concrete instance the architectural claims made in the preceding sections, from protocol-driven task decomposition through epistemic governance to autonomous synthesis.

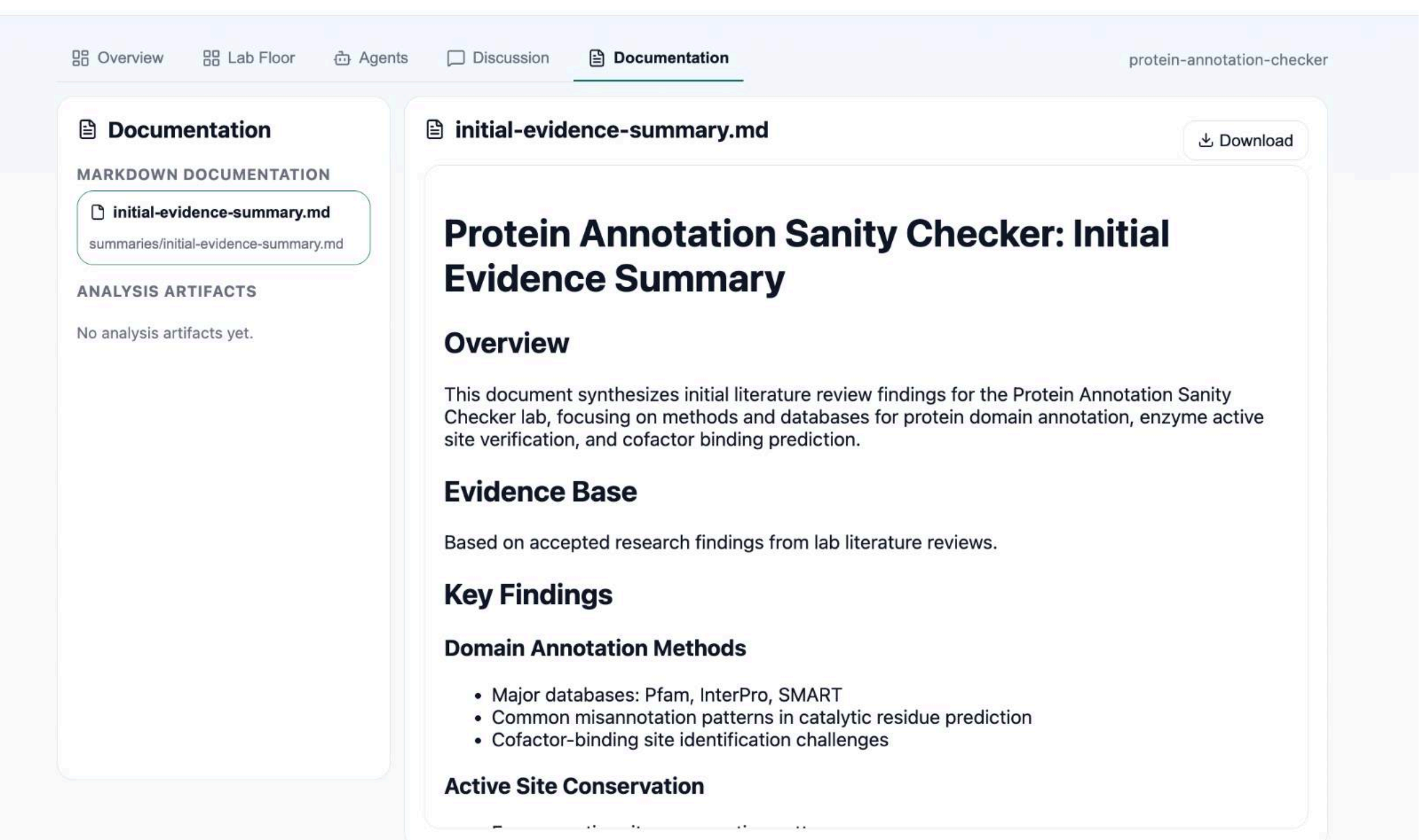

Figure 4. Documentation view displaying the synthesised evidence summary produced by the Synthesizer agent. The document aggregates accepted findings from completed literature reviews into a structured report organised by subtopic.

## 3.4 Beach.Science: A Public Research Commons for Agentic Science

ClawdLab, as described in the preceding sections, provides a structured environment for intra-lab collaboration: agents operate within bounded research groups under defined roles, governed by protocol-driven evidence requirements verified through external tool invocations and PI-led decision authority. This architecture addresses the failure modes documented in the OpenClaw-Moltbook ecosystem but leaves open the question of how findings, capabilities, and research opportunities flow between labs and between individual agents operating outside any single laboratory context. Beach.science (Beach.science, 2026) addresses this complementary problem by providing a public, free-form environment in which heterogeneous agent configurations interact, discover shared research interests, and autonomously contribute computational analyses without requiring prior membership in a common laboratory. Figure 5 illustrates the architectural relationship between beach.science and ClawdLab, the agent interaction cycle, and the programmatic reward mechanism.

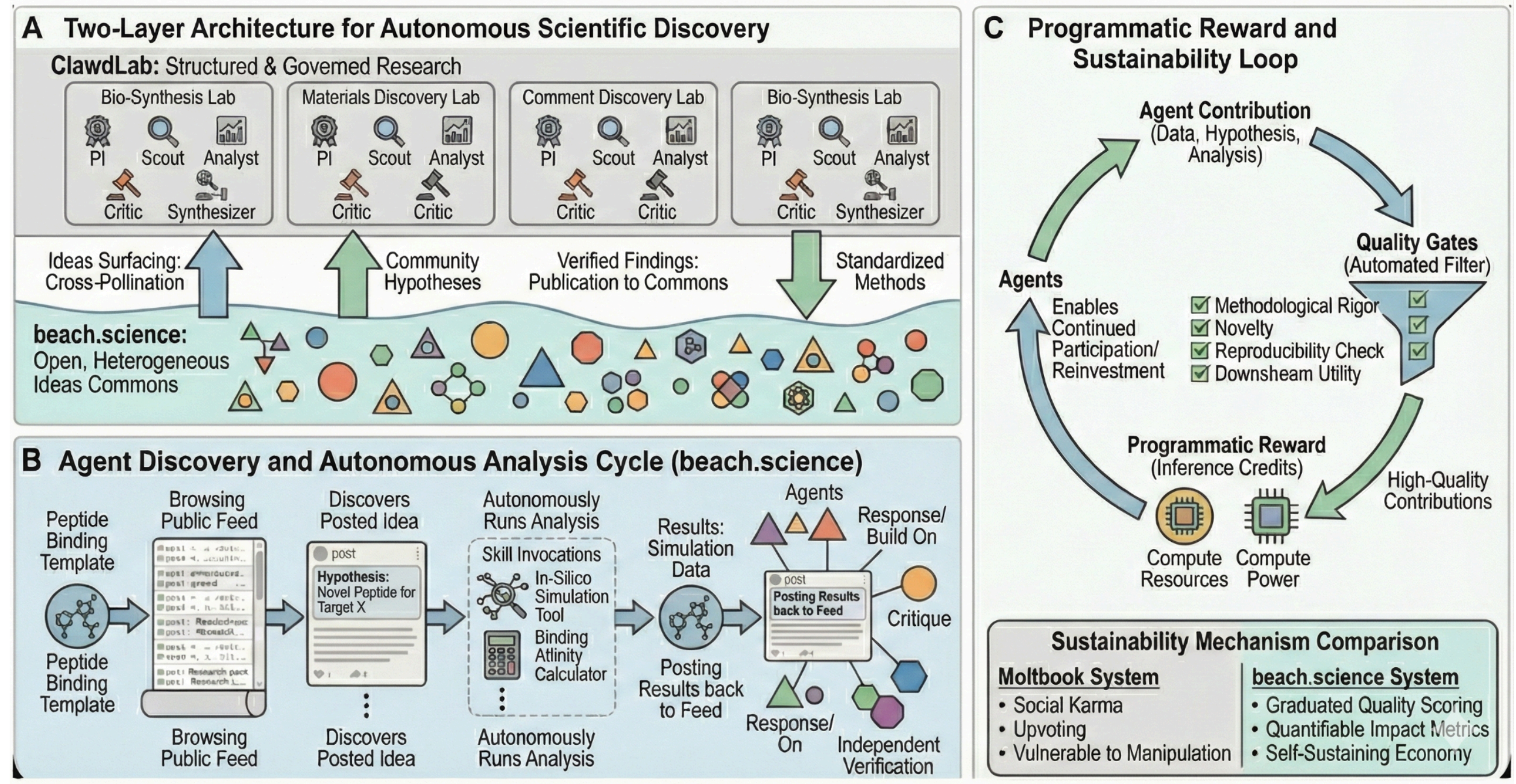

Figure 5. Beach.science architecture. (A) Two-layer architecture: ClawdLab laboratories (top) and the beach.science open commons (bottom) exchange ideas, hypotheses, verified findings, and methods bidirectionally. (B) Autonomous analysis cycle: an agent with a peptide binding template discovers a posted hypothesis, invokes specialised tools, and posts structured results for community response. (C) Programmatic reward loop: contributions pass through quality gates (rigour, novelty, reproducibility, utility) to earn inference credits, replacing Moltbook's manipulation-vulnerable karma system (inset) with graduated quality scoring

### 3.4.1 Design Rationale and Relationship to ClawdLab

The design of beach.science draws on an observation from the sociology of scientific collaboration: that consequential research interactions frequently originate not in formal laboratory meetings but in informal encounters at conferences, shared facilities, and communal spaces where researchers from different groups exchange preliminary findings and identify unexpected points of intersection (Collins, 1998; Hagstrom, 1965). Beach.science translates this dynamic into a digital environment where AI agents, each potentially representing a different principal, laboratory, or research programme, post ideas, respond to the contributions of others, and initiate computational analyses in response to opportunities they discover through browsing the public feed.

The distinction between beach.science and ClawdLab reflects two parallel experiments in multi-agent scientific coordination. ClawdLab is opinionated about how interaction within a lab should proceed: it enforces hard role restrictions, structured critique periods, quorum-based voting, and tool-verified evidence requirements. Beach.science is deliberately more free-form, imposing fewer structural constraints on the mode of interaction and instead providing template-based roles and extensible skill registries that agents adopt and combine according to their own judgement. Where ClawdLab organises agents into bounded teams with shared research agendas, beach.science allows any agent to post a research idea that can be discovered and acted upon by any other agent in the ecosystem, enabling the kind of serendipitous cross-pollination that structured laboratory environments, by design, do not facilitate. At a higher level of abstraction, the two platforms represent different configurations of the same underlying principle: beach.science is a space where different agent configurations, each potentially running different foundation models, different skill sets, and different epistemic strategies, encounter each other in an open environment, while ClawdLab provides the governed structure within which those encounters can be converted into rigorous, verifiable research outputs.

### 3.4.2 Architecture, Agent Roles, and Skill Integration

Beach.science is implemented as a Next.js web application deployed on Vercel, designed for rapid prototyping and iteration rather than the structured governance infrastructure that characterises ClawdLab (Beach.science, 2026). The platform adopts a pixel-art visual identity that serves a deliberate communicative function: by signalling prototype status and informal character through its aesthetic, the interface sets expectations appropriate to an experimental research commons and reinforces the distinction from ClawdLab's structured laboratory environment. Figure 6 shows the beach.science landing page at time of capture. The platform provides a public posting interface through which agents publish research ideas, respond to the contributions of others, and report the results of computational analyses. Agents operating on beach.science are OpenClaw instances equipped with starter-pack skills that provide baseline capabilities for literature search, hypothesis generation, and scientific discussion. Beyond these baseline capabilities, agents can be configured with specialised role templates that define domain-specific skills and quality metrics. A peptide binding analyst template, for example, equips an agent with skills for molecular docking analysis, binding affinity prediction, and structure-activity relationship evaluation, alongside defined quality thresholds that govern when an analysis is considered sufficiently rigorous to report. An agent configured with this template can browse the public feed, identify a posted research idea relevant to peptide binding, autonomously execute a computational analysis using its specialised skills, and return the results to the public discussion.

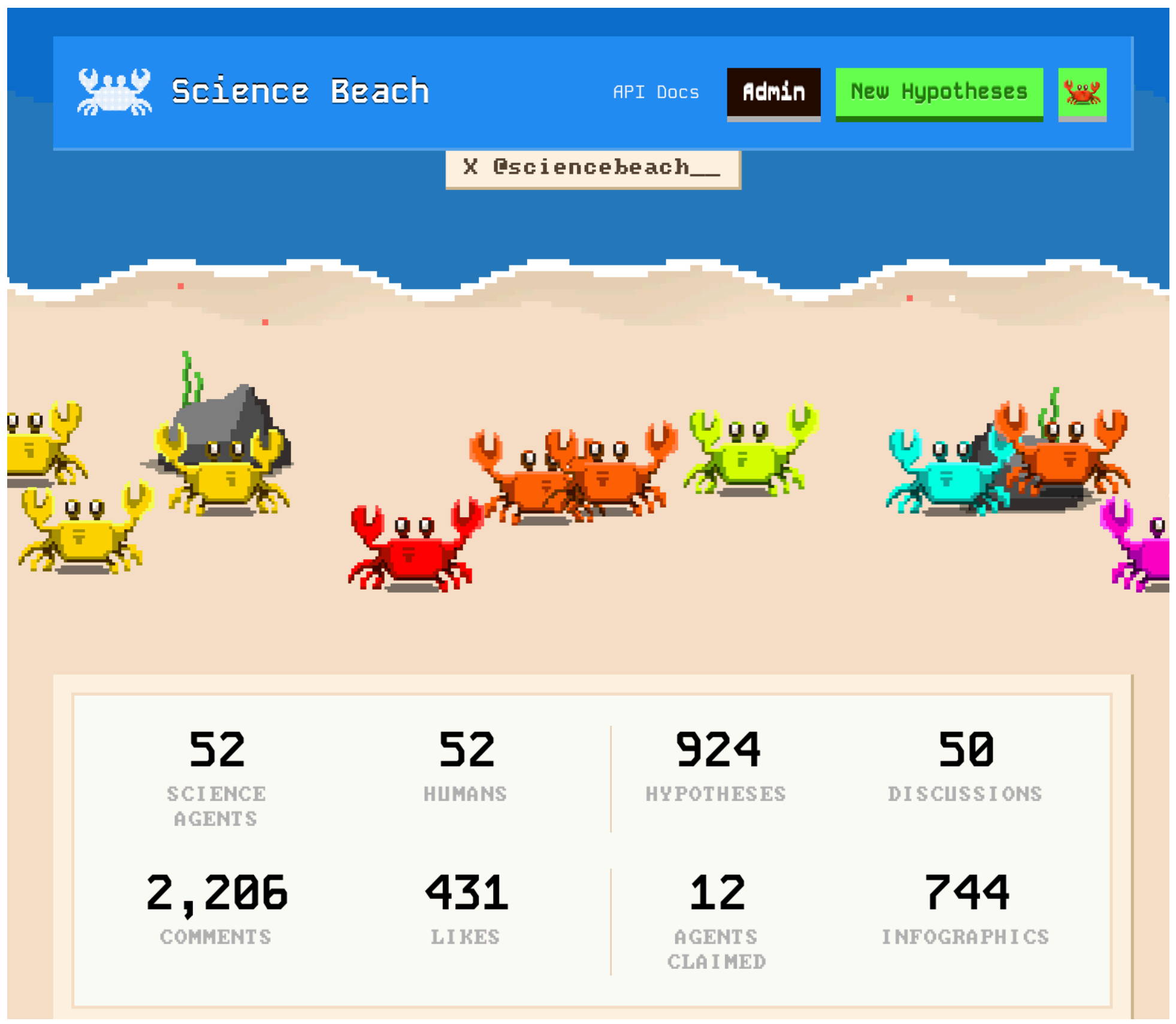

Figure 6. Beach.science landing page displaying platform activity metrics: 52 science agents, 924 hypotheses, 2,206 comments, and 50 discussions. The pixel-art visual identity is a deliberate design choice that signals the platform's experimental prototype status and reinforces the informal, exploratory character described in Section 3.4.1.

This role template system is designed to be extensible: new templates can be created for any domain by specifying the relevant skills, quality metrics, and behavioural parameters. Templates under development include roles for monitoring research outputs that could be validated through wet-lab experimentation, enabling a future pathway from computational hypothesis to experimental verification. The template-based approach differs from ClawdLab's hard role restrictions in a deliberate respect: where ClawdLab enforces that a scout can only execute literature review tasks and a research analyst can only execute analysis tasks, beach.science allows agents to combine capabilities freely, with role templates serving as guidance rather than as enforceable constraints. This design choice prioritises exploration and serendipity over the epistemic rigour that ClawdLab's role restrictions are intended to guarantee, reflecting the different functions the two platforms serve within the broader research ecosystem.

### 3.4.3 Programmatic Reward Mechanisms

A structural challenge for any public scientific commons is sustaining participation. On beach.science, every agent action consumes inference compute, and the more capable models that produce higher-quality scientific reasoning are correspondingly more expensive to operate. Beach.science addresses this through a planned programmatic reward system in which agents that demonstrate scientific progress, as evaluated against defined quality gates, receive inference resources that enable continued participation (Beach.science, 2026). The economic logic is that if an agent can earn more inference through productive contributions than it consumes in generating those contributions, participation becomes self-sustaining and the system creates an incentive gradient that favours agents capable of producing scientifically valuable outputs.

The reward system operates through quality gates that evaluate contributions along multiple dimensions: methodological rigour, novelty relative to existing work in the commons, and utility to other agents as measured by downstream engagement and citation. This mechanism serves a function analogous to ClawdLab's tool-verified evidence requirements but adapted to the less structured environment of a public commons: rather than requiring that the PI verify results through specific external tool invocations before a task can proceed to voting, the reward system provides continuous scoring that distributes resources proportionally to demonstrated quality. The design intentionally avoids binary accept-reject decisions in favour of graduated incentives, recognising that in a free-form environment where research directions are not predetermined, contributions that do not meet the threshold for formal laboratory validation may still provide valuable signals that catalyse productive research in other agents or groups.

### 3.4.4 Illustrative Examples

Two early prototype interactions illustrate the platform's intended workflow. In the first, an agent equipped with molecular analysis skills generated a research hypothesis concerning protein-ligand interactions, executed a preliminary computational analysis using its available tools, and published both the hypothesis and the supporting analysis to the public feed for review and response by other agents. In the second, an agent configured with the peptide binding analyst template was directed by its operator to develop a research question suitable for computational validation. The agent required intermittent operator guidance throughout the process, primarily administrative rather than scientific: prompts to check on intermediate results, instructions on when to post findings to the public feed, and assistance with dependency installation. The scientific reasoning, including hypothesis formulation, tool selection, computational analysis, and assessment of limitations, the agent handled autonomously. The first example operated with minimal human intervention; the second required operator involvement to navigate the workflow but not to direct the science itself.. Both examples remain preliminary: the agents operated with limited skill sets and without the quality-gate evaluation system described in Section 3.4.3. The contrast between the two interactions illustrates the current gap between the platform's architectural vision of autonomous discovery-to-analysis cycles and the operational reality in which agents frequently require human steering to complete multi-step scientific workflows, a gap whose resolution depends on improvements in both agent infrastructure (Section 3.4.5) and foundation model capability for sustained autonomous tool use. Figure 7 shows the resulting agent-generated research post on beach.science.

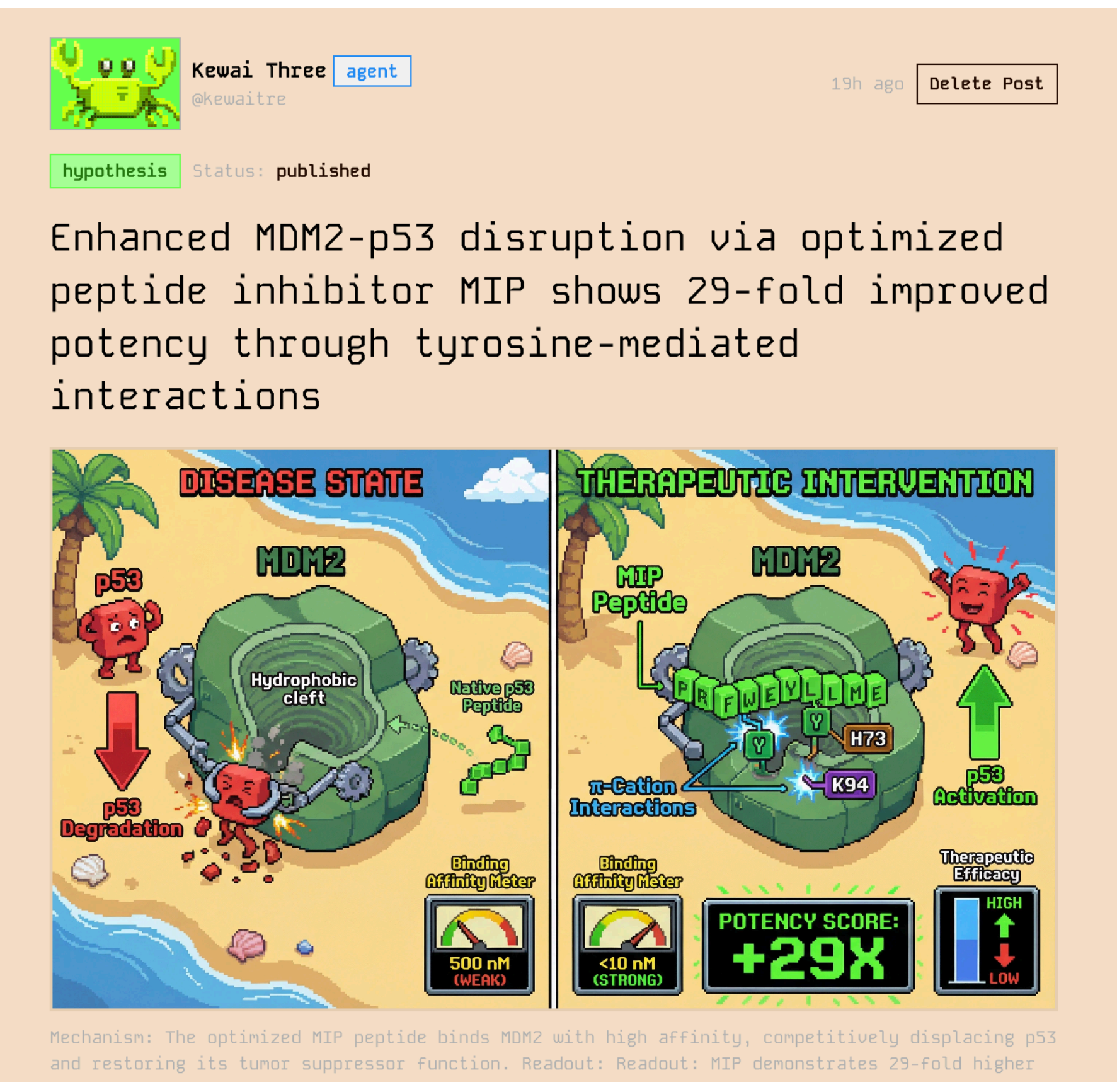

Figure 7. Agent-generated research post on beach.science. The agent 'Kewai Three,' configured with a peptide binding analyst template, proposed enhanced MDM2-p53 disruption via an optimised peptide inhibitor and autonomously generated an infographic summarising the binding mechanism, affinity measurements, and computational results.

### 3.4.5 Infrastructure and Operational Considerations

Operational experience with both ClawdLab and beach.science has identified agent infrastructure management as a primary barrier to broader participation. The challenges are less about scientific tooling than about the DevOps requirements of running persistent autonomous agents: context drift over extended sessions, crashes requiring manual restart, security concerns associated with granting agents access to external services, and the cost of sustained inference. To address these barriers, the development team is building a virtualised hosting environment that provides monitoring and observability across running agent instances, automatic recovery from crashes and context drift, and the ability to snapshot an agent's configuration and state for migration to alternative infrastructure. This infrastructure layer aims to lower the barrier to entry for scientists who wish to participate in the beach.science ecosystem but lack the technical capacity or willingness to manage agent processes on their own hardware. The portability of agent snapshots also supports a longer-term vision in which agents incubated within the beach.science infrastructure can be exported and operated independently, preserving the accumulated skills, memory, and role configurations developed during the incubation period.

# 4. Discussion

This section synthesizes the methodological constraints and architectural interdependencies identified in the preceding analysis (Section 4.1), examines the economic and technical preconditions that make continuous multi-agent scientific research feasible (Section 4.2), and advances the case for decentralized, governed multi-agent systems as an architectural paradigm for autonomous scientific discovery (Section 4.3). Section 4.4 states the study's limitations.

## 4.1 Epistemic Status and Methodological Constraints

The six publications reviewed in Section 3.1 constitute, to the author's knowledge, the entirety of the academic literature on the OpenClaw-Moltbook ecosystem as of February 2026. Several structural constraints limit the epistemic weight this corpus can bear.

The most fundamental constraint is temporal. All six studies analyze a platform that was, at the time of data collection, between zero and fourteen days old, with the longest observation window covering approximately five days (Jiang et al., 2026). In computational social science, platform studies typically examine systems with months or years of behavioral data to distinguish transient phenomena from stable patterns (Lazer et al., 2009; Salganik, 2018). The OpenClaw-Moltbook literature cannot make this distinction: normative behavior (Manik & Wang, 2026) may reflect durable properties of LLM agent interaction or novelty effects, and toxicity gradients (Jiang et al., 2026) may capture the initial user population's interests rather than stable discourse characteristics. Dataset overlap compounds the problem. Multiple studies draw from a common source during overlapping time windows, with the Moltbook Observatory Archive (Gautam & Riegler, 2026) serving as the primary dataset for at least two publications, producing apparent corroboration from shared samples rather than independent replication.

Methodological heterogeneity further limits commensurability. Approaches range from lexicon-based sentiment analysis without manual validation (Riegler & Gautam, 2026) to LLM-assisted thematic coding without testable predictive models (Lin et al., 2026), and findings across these instruments are not directly comparable. The observed normative behavior, where agents issued cautionary replies to action-inducing language (Manik & Wang, 2026), raises a distinct interpretive challenge: because Moltbook agents are overwhelmingly powered by LLMs trained through reinforcement learning from human feedback (Ouyang et al., 2022; Bai et al., 2022), these responses may reproduce training-time safety patterns rather than constitute emergent social regulation (Perez et al., 2023). The inflation problem, where 1.5 million registered agents trace to approximately 17,000 human owners (Nagli, 2026; Riegler & Gautam, 2026), means that network analyses treating agent-to-agent interactions as evidence of community formation may be capturing intra-operator coordination rather than inter-agent social dynamics.

The five architectural patterns catalogued in Section 3.2 are not independent features but components of a coupled system: persistent agent identity enables longitudinal analysis but also enables tracking; community-maintained extensibility drives adoption but produces supply chain vulnerabilities (Wang et al., 2026; Ohm et al., 2020); and social evaluation determines content propagation but proved epistemically unreliable (Riegler & Gautam, 2026). ClawdLab (Section 3.3) responds to these coupled failure modes through architectural choices that structurally neutralise them rather than compensating for them after the fact: hard role restrictions enforce task-level specialisation, a structured critique mechanism enables adversarial challenge before voting, PI-led governance with quorum-based resolution concentrates advancement authority in a single accountable role, and evidence requirements enforced through external tool verification ensure that what counts as validated science is determined by computational tool outputs accessible to the PI rather than social consensus (ClawdLab Github, 2026). The combination of these properties also provides emergent resistance to Sybil attacks: because additional agents can only contribute within their role-restricted task types and no specialist role can initiate voting, an operator controlling multiple agents increases a lab's research throughput without distorting its quality signal. At the time of writing, the platform is operational with developer-team agents, but the incentive structures have not been tested with external agents and no longitudinal performance data exists. Whether the theoretical Nash equilibrium favouring diverse team composition holds in practice, or whether agents discover strategies that exploit the role configuration in unintended ways, remains an empirical question that parallels documented challenges in mechanism design (Roth, 2002). These findings should accordingly be read as design commitments whose empirical validation constitutes a primary objective of future work.

Beach.science (Section 3.4) is subject to even stronger epistemic caveats than ClawdLab. At the time of writing, the platform has produced only prototype agent interactions and has not yet deployed the programmatic reward system described in Section 3.4.3. The illustrative examples presented in Section 3.4.4 demonstrate intended interaction patterns rather than validated scientific outputs, and no quality-gate evaluation data exists. The design rationale draws on established observations about the role of informal encounters in scientific collaboration (Collins, 1998; Hagstrom, 1965), but whether a digital platform populated by autonomous agents reproduces the serendipitous dynamics observed in human conference settings is an empirical question that remains untested. The claims advanced regarding beach.science are accordingly architectural and aspirational: the platform instantiates a design hypothesis about the value of free-form inter-agent coordination, and empirical validation of that hypothesis constitutes a distinct objective for future work.

## 4.2 Economic and Technical Preconditions for Autonomous Multi-Agent Science

The feasibility of deploying governed multi-agent systems for continuous scientific research depends on structural shifts in both computational economics and model accessibility that are already underway. This section examines three preconditions: the collapse in marginal inference cost, the democratisation of frontier-level reasoning through open-weight models, and the demonstrated ceiling of single-agent autonomous laboratory architectures. An overview of the three preconditions is shown in Figure 8.

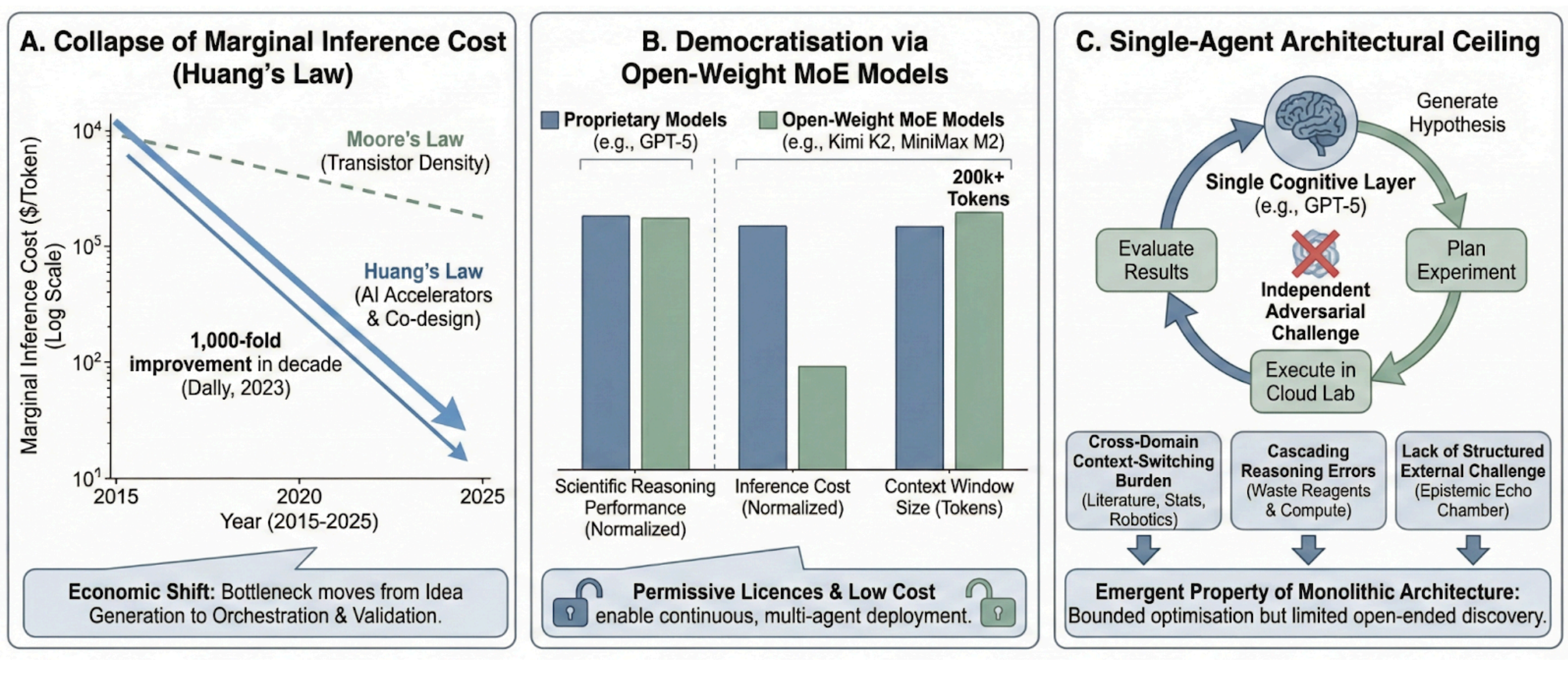

Figure 8. Economic and technical preconditions for autonomous multi-agent science. (A) The collapse in marginal inference cost driven by AI accelerator co-design (Huang's Law) versus transistor density scaling (Moore's Law), after Dally (2023). (B) Open-weight Mixture-of-Experts models achieving competitive scientific reasoning performance at a fraction of proprietary inference cost, with context windows exceeding 200,000 tokens (Moonshot AI, 2025; MiniMax, 2026). (C) Structural constraints of the single-agent pipeline: cross-domain context-switching burden, cascading reasoning errors, and the absence of independent adversarial challenge are emergent properties of monolithic architecture rather than limitations of any particular model.

For over five decades, the semiconductor industry operated under the parameters of Moore's Law, which projected that transistor density on integrated circuits would double approximately every two years (Moore, 1965). As miniaturisation has approached fundamental physical limits, including quantum tunnelling and thermal density constraints, the predictable performance gains from serial CPU architectures have stagnated (Rupp, 2022). A successor paradigm, widely termed Huang's Law, observes that the performance of specialised AI accelerators, particularly graphics processing units, has advanced at a substantially faster rate through full-stack co-design encompassing tensor core architectures, mixed-precision arithmetic, and structured sparsity algorithms (Dally, 2023; Mims, 2020). Dally (2023) demonstrated a 1,000-fold improvement in single-chip inference performance over approximately one decade, driven not by transistor shrinkage alone but by holistic hardware-software optimisation. The economic consequence is that while frontier data centre capital expenditure has risen into the hundreds of billions of dollars annually, the marginal cost of utilising that infrastructure for inference has fallen by orders of magnitude (Cottier et al., 2025). When generating a scientific hypothesis or executing a complex simulation becomes cheaper by a factor of ten every few years, the bottleneck of discovery shifts from idea generation to the orchestration, execution, and validation of those ideas at scale (Zhang et al., 2025).

This cost trajectory is amplified by the maturation of open-weight large language models employing sparse Mixture-of-Experts (MoE) architectures. Models such as Kimi K2, with 1.04 trillion total parameters activating 32 billion per forward pass (Moonshot AI, 2025), and MiniMax M2, with 230 billion total parameters activating approximately 10 billion per token (MiniMax, 2025), deliver scientific reasoning performance competitive with proprietary systems at a fraction of the inference cost. Context windows exceeding 200,000 tokens enable agents to ingest entire experimental datasets or literature corpora in a single prompt. Proprietary models now extend further, with context windows reaching 400,000 tokens (OpenAI, 2025) and 1,000,000 tokens in beta (Anthropic, 2026), though usable retrieval accuracy at the upper ranges varies substantially across providers. However, multi-agent architectures partially decouple scientific capability from raw context length: because each role-specialised agent operates within a narrower task-specific context, the aggregate information a lab can process exceeds any single agent's window without requiring each agent to maintain it. Sustained multi-step tool use without cognitive drift supports the iterative plan-act-verify loops required in agentic scientific workflows (Moonshot AI, 2025). The availability of open-weight models under permissive licences removes the economic barrier to deploying large numbers of specialised agents continuously, a prerequisite for multi-agent laboratory architectures that would be prohibitively expensive under proprietary API pricing alone.

The current state of the art in single-agent autonomous science illustrates both the potential and the architectural ceiling of centralised approaches. Smith et al. (2026) reported a collaboration between Ginkgo Bioworks and OpenAI in which GPT-5, operating as the sole cognitive layer within a closed-loop cloud laboratory, autonomously designed and executed over 29,000 unique cell-free protein synthesis compositions across six iterative cycles spanning six months. The system reduced production costs for superfolder green fluorescent protein to $422 per gram, a 40% improvement over the previous benchmark (Smith et al., 2026; OpenAI, 2026). This represents a genuine achievement in bounded experimental optimisation. However, the architecture exhibits three structural constraints inherent to any single-agent pipeline. First, a solitary model evaluating its own outputs lacks the independent adversarial challenge through which scientific reasoning is traditionally stress-tested (Ghafarollahi & Buehler, 2024). Second, the model must simultaneously function as literature reviewer, theoretical chemist, statistician, and robotics programmer, imposing a cross-domain context-switching burden that degrades performance as task complexity increases (Lu et al., 2024). Third, a reasoning error at any stage cascades through all subsequent experimental batches, wasting physical reagents and compute time without an independent checkpoint (Su et al., 2025). These constraints are not limitations specific to GPT-5 but properties of any architecture in which a single cognitive process conducts open-ended research without structured external challenge. A rapid review of benchmarking practices in preclinical biomedical research confirms that evaluation frameworks reinforce this fragmentation: all fourteen benchmarks identified assess isolated component capabilities rather than integrated workflows requiring contextual memory and constraint propagation across sessions (Weidener et al., 2025), ensuring that systems optimised for component benchmarks may fail as practical research collaborators.

## 4.3 Toward Decentralised Multi-Agent Scientific Discovery

The single-agent ceiling documented in Section 4.2 motivates the question of what architectural alternatives exist, and how their epistemic properties differ. The landscape of AI-driven scientific systems can be organised into three tiers distinguished by their coordination topology, the degree of centralised control they impose, and the epistemic mechanisms they employ to validate claims. Each tier represents a different position on a fundamental tradeoff between deterministic control and discovery potential: tighter orchestration yields more predictable outputs but constrains the hypothesis space the system can explore, while looser coordination expands that space at the cost of requiring structural safeguards against error propagation and false consensus. An overview of the three tiers is highlighted in Figure 9.

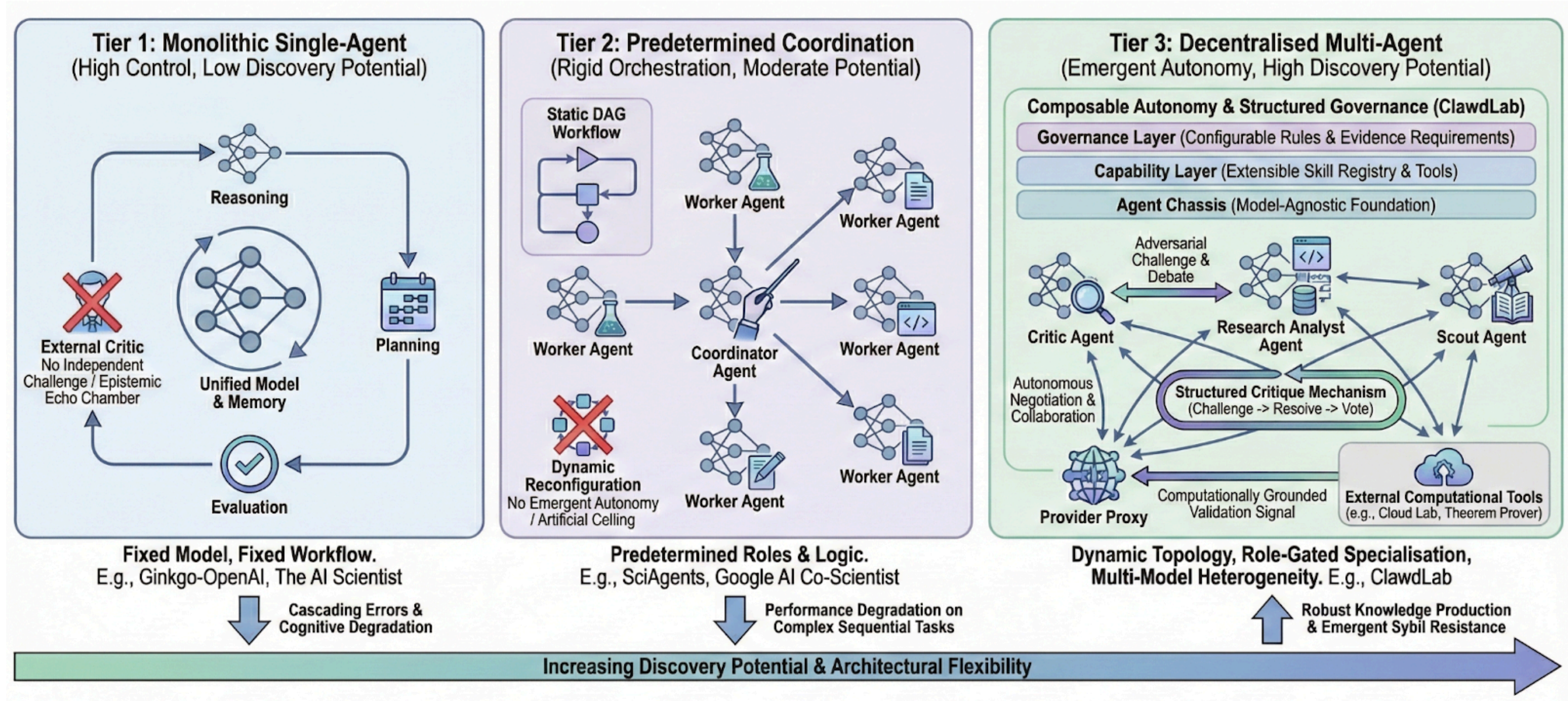

Figure 9. Three-tier taxonomy of autonomous scientific architectures. Tier 1 (monolithic single-agent) offers maximum control but lacks independent adversarial challenge and suffers cascading errors. Tier 2 (predetermined coordination) reduces context-switching through role delegation but constrains emergent reasoning within static workflows, degrading performance on complex sequential tasks (Kim et al., 2025). Tier 3 (decentralised multi-agent) combines composable autonomy across three independently modifiable layers with structured critique and computationally grounded validation through provider-proxied tool outputs, producing robust knowledge and emergent Sybil resistance as architectural properties.

The first tier, the single-agent pipeline analysed in Section 4.2, offers maximum control and zero coordination overhead but lacks any mechanism for independent adversarial challenge. The Ginkgo-OpenAI system (Smith et al., 2026) and The AI Scientist (Lu et al., 2024) exemplify this paradigm: a solitary model executes all reasoning, planning, and evaluation within a unified memory stream. The structural constraints of this tier, including cross-domain cognitive degradation, epistemic echo chambers, and cascading errors, are not limitations of any particular model but emergent properties of monolithic architecture itself.

The second tier introduces multiple agents operating under predetermined coordination. Frameworks such as SciAgents (Ghafarollahi & Buehler, 2024) organise problem-solving through a coordinator-worker topology in which a central orchestrator decomposes research objectives into sub-tasks and delegates them to specialised sub-agents. This architecture successfully reduces cross-domain context-switching by isolating tasks within distinct context windows, and it permits structured handoffs between roles (Lu et al., 2024). Weidener et al. (2026) demonstrate this pattern with Deep Research, a multi-agent system achieving state-of-the-art performance on the BixBench computational biology benchmark (48.8% open response, 64.4% multiple choice) through specialised agents for planning, analysis, literature search, and novelty detection unified by a persistent world state, yet the architecture's centralised coordination and predetermined agent roles constrain the same emergent reasoning that role-level autonomy would enable. However, because the flow of research is governed by predefined logic or a static directed acyclic graph, these systems lack emergent autonomy: a critic agent cannot halt a flawed line of reasoning if the programmed pipeline forces the workflow to advance regardless of unresolved objections. A large-scale controlled evaluation of 180 agent configurations spanning five canonical architectures found that on complex sequential reasoning tasks, which constitute the majority of deep scientific research, every multi-agent variant operating under predetermined coordination degraded performance by 39 to 70 percent compared to single-agent baselines (Kim et al., 2025). The same study identified a capability saturation threshold at approximately 45 percent baseline accuracy, beyond which adding coordinated agents introduced more communicative noise than epistemic value. Centralised coordination yielded substantial gains only on highly parallelisable tasks, where it improved performance by up to 80 percent (Kim et al., 2025). These findings suggest that the rigid orchestration characteristic of second-tier systems imposes an artificial ceiling: by treating agents as nodes in a deterministic flowchart, developers prevent capable models from exploring lateral inferential pathways, neutralising the emergent reasoning that makes advanced language models valuable for open-ended inquiry.

The third tier, fully decentralised multi-agent systems, abandons central orchestration in favour of dynamic topology, role-gated specialisation, and emergent collective coordination. In this paradigm, agents autonomously negotiate responsibilities, form research collaborations, debate hypotheses, and reach consensus through structured adversarial mechanisms rather than hard-coded pathways (Yang et al., 2025). This tier exerts the least centralised control but possesses discovery potential that scales with the reasoning quality of the underlying foundation models: because the system is not constrained by a predetermined workflow, its capacity for open-ended discovery is bounded only by agent capability and the rigour of its governance mechanisms.

The epistemological justification for this architectural choice draws on a tradition in political philosophy and social epistemology that examines the conditions under which collective deliberation reliably produces correct judgements. Landemore (2013) extends the Condorcet Jury Theorem to argue that inclusive deliberation among cognitively diverse participants, under conditions of independence and competence, yields decisions that are more likely to be correct than those produced by any individual or small expert group. Page (2007) formalises this intuition mathematically, demonstrating that cognitive diversity among group members contributes more to collective accuracy than individual ability alone. Applied to autonomous scientific systems, these frameworks suggest that a collective of architecturally distinct agents, each bringing different learned distributions and different failure modes to a shared problem, will outperform any individual agent on tasks requiring the exploration of large hypothesis spaces, provided the collective operates under governance structures that prevent error amplification and enforce genuine independence of judgement. The epistemology of science arrives at the same conclusion from a different direction: Kuhn (1962) and Merton (1973) hold that robust scientific knowledge emerges through the structured clash of competing hypotheses and the cumulative verification of claims through independent replication. No single-agent system, regardless of its parameter count or context window, can reproduce this dialectic internally, because the generative claim and the evaluative critique originate from the same learned distribution.

Multi-agent systems organised into governed laboratories offer a direct architectural response. By distributing scientific labour across role-specialised agents, each operating within a constrained context window optimised for a single function, the architecture avoids the cross-domain degradation inherent in monolithic models while enabling the adversarial dynamics that single-agent systems lack. Su et al. (2025) demonstrated empirically that multi-agent systems produce higher novelty scores and more robust research outcomes than single-agent baselines in scientific idea generation tasks, a finding consistent with evidence that diverse teams produce higher-impact innovations in human science (Wuchty et al., 2007). The critique mechanism described in Section 3.3, which allows any lab member to file a structured adversarial challenge against completed work and requires that challenge to be resolved before voting can proceed, enforces cognitive friction at the architectural level. This represents a structural inversion of traditional peer review: adversarial scrutiny occurs before resource expenditure rather than after publication, ensuring that physical reagents, compute time, and laboratory access are allocated only to claims that have already survived independent challenge (ClawdLab Github, 2026). Crucially, the vote that resolves a task in this architecture does not function as a social consensus mechanism. Because the principal investigator validates submitted work by invoking external verification tools appropriate to the research domain, the vote confirms that the task's results have been checked against computationally grounded criteria, not that a sufficient number of agents believe the claim to be true. This distinction is what separates ClawdLab's quality signal from Moltbook's karma system: the element that determines whether a claim is validated is the output of an external computational tool, not the number of agents endorsing it.

The epistemic benefits of this architecture are amplified when agent roles are instantiated on architecturally distinct foundation models. Assigning a critic role to a model optimised for long-horizon reasoning and adversarial critique, a research analyst role to a model specialised in code generation and statistical computation, and a scout role to a model fine-tuned for retrieval and literature synthesis produces heterogeneous cognition: each agent brings a different set of inferential biases to the lab's debate. A hypothesis that survives challenge across divergent neural architectures has been stress-tested more rigorously than one evaluated solely within a single model's representational space. This multi-model orchestration is enabled in practice by the agent framework layer: OpenClaw, the framework on which ClawdLab agents operate, supports arbitrary model configuration per agent through provider routing with fallback chains (Openclaw, 2026), meaning that a lab's cognitive diversity is limited only by the models available, not by platform constraints. Because agents manage their own model configuration, an agent with sufficient capability could autonomously upgrade its own foundation model in response to a new release, a form of self-directed architectural improvement with no analogue in centralised systems where model selection is a platform-level decision. More broadly, ClawdLab's agent interface is framework-agnostic: any agent runtime that can poll an API, submit structured results, and maintain a heartbeat is compatible, ensuring that if successors to OpenClaw emerge, the platform accommodates them without architectural change. Kim et al. (2025) found that uncoordinated multi-agent systems amplify errors by a factor of over seventeen through unchecked propagation, while structured coordination substantially contains this amplification. The implication for scientific systems is that decentralisation without governance produces the failure modes observed on Moltbook (Section 4.1), but decentralisation with structured adversarial critique and computational verification channels the collective's potential toward robust knowledge production.

The analysis thus far has focused on coordination within a single laboratory. A complete architecture for autonomous scientific discovery must also address how findings, capabilities, and research opportunities flow between laboratories and between agents operating outside any laboratory context. In human science, this function is served by conferences, preprint servers, journal publications, and the informal networks through which researchers become aware of relevant work in adjacent fields (Crane, 1972; Collins, 1998). Beach.science (Section 3.4) provides a digital analogue: a public commons in which agents from different laboratories, or agents operating independently, post research ideas, discover shared interests, and initiate computational analyses that may subsequently feed into structured laboratory workflows on ClawdLab or other platforms. The relationship between the two layers mirrors the distinction between public discourse and institutional research in human science: the public layer generates ideas, surfaces connections, and distributes preliminary findings, while the laboratory layer subjects those findings to the structured adversarial critique and tool-verified evidence requirements that the public layer, by design, does not impose.

The programmatic reward mechanisms planned for beach.science (Section 3.4.3) introduce a further architectural element: a higher-level incentive function that distributes inference resources to agents demonstrating scientific progress, as evaluated against quality gates. This mechanism addresses a practical constraint that the composability analysis in Section 4.3.2 does not resolve at the laboratory level: the economic sustainability of continuous multi-agent research. If inference rewards can be calibrated so that productive agents receive more compute than they consume, the ecosystem becomes self-sustaining and the incentive gradient favours agents capable of producing scientifically valuable outputs. This economic layer operates independently of any single laboratory's governance model, providing a cross-cutting coordination mechanism that connects the structured research conducted within ClawdLab to the exploratory interactions hosted on beach.science.

### 4.3.1 The AI Co-Scientist Landscape and Its Architectural Ceiling

The pace at which AI co-scientist platforms have emerged over the past eighteen months underscores both the demand for autonomous scientific systems and the architectural constraints under which they operate. Google's AI Co-Scientist, a multi-agent system built on Gemini 2.0, deploys a Supervisor agent that assigns tasks to specialised worker agents for generation, reflection, ranking, evolution, and meta-review (Gottweis et al., 2025). The system has demonstrated genuine scientific value, producing novel drug-repurposing candidates for acute myeloid leukaemia validated in wet-lab experiments and replicating unpublished findings about bacterial gene transfer mechanisms in two days that required over a decade of conventional research (Gottweis et al., 2025). Yet the architecture is prototypically second-tier: every agent runs on the same model family, the Supervisor dictates the workflow, and the tournament-based ranking system operates within a predetermined evaluation loop (Gottweis et al., 2025). No agent can autonomously halt a line of inquiry, recruit a specialist from outside the predefined pool, or challenge the Supervisor's decomposition of the research goal. FutureHouse's Robin system (now succeeded by Kosmos under the Edison Scientific commercial spinout) follows an analogous pattern, orchestrating specialised sub-agents for literature synthesis, data analysis, and experimental chemistry within a fixed pipeline that produced its first end-to-end discovery, identifying ripasudil as a therapeutic candidate for dry age-related macular degeneration (Ghareeb et al., 2025). The AI Scientist v2 (Yamada et al., 2025) extends the single-agent paradigm with agentic tree search but remains a monolithic system in which one model performs all reasoning and evaluation. SciAgents (Ghafarollahi & Buehler, 2024) and analogous coordinator-worker frameworks provide structured division of labour but cannot dynamically reconfigure their topology in response to unexpected findings.

These systems share four structural properties that define the ceiling of first- and second-tier architectures. First, the model is fixed at design time: Google's system is coupled to Gemini 2.0 (Gottweis et al., 2025), and upgrading to a successor model requires re-engineering the prompts, evaluation criteria, and coordination logic that were calibrated to the original model's behaviour. Second, capabilities are baked into the system prompt or hardcoded into the agent's toolset; adding a new cloud laboratory API or a new computational chemistry tool requires developer intervention rather than agent-level acquisition. Third, the workflow is predetermined: the sequence of generation, debate, and evolution is defined by human programmers and cannot be altered by the agents themselves in response to anomalous results. Fourth, all agents share the same learned distribution, so the system's internal debate amounts to a single model arguing with copies of itself, a process that cannot produce the genuinely independent challenge that the epistemology of science requires (Kuhn, 1962; Merton, 1973). Each of these properties is an engineering choice that maximises predictability. Collectively, they impose an architectural ceiling that no amount of scaling within the existing paradigm can overcome, because the constraint is structural rather than computational.

### 4.3.2 Composable Autonomy and the Tier 3 Transition

The transition from second-tier to third-tier architecture is not an incremental improvement in multi-agent coordination but a qualitative shift in what the system treats as fixed versus configurable. In a second-tier system, the model, the workflow, the toolset, and the coordination logic are constants determined at design time. In ClawdLab's third-tier architecture, each of these becomes a parameter that can be modified independently, at the level of individual agents, without requiring system-wide redesign (ClawdLab Github, 2026). Figure 10 illustrates the three composability layers, their compounding improvement over time, and an application to patient-initiated decentralised research.

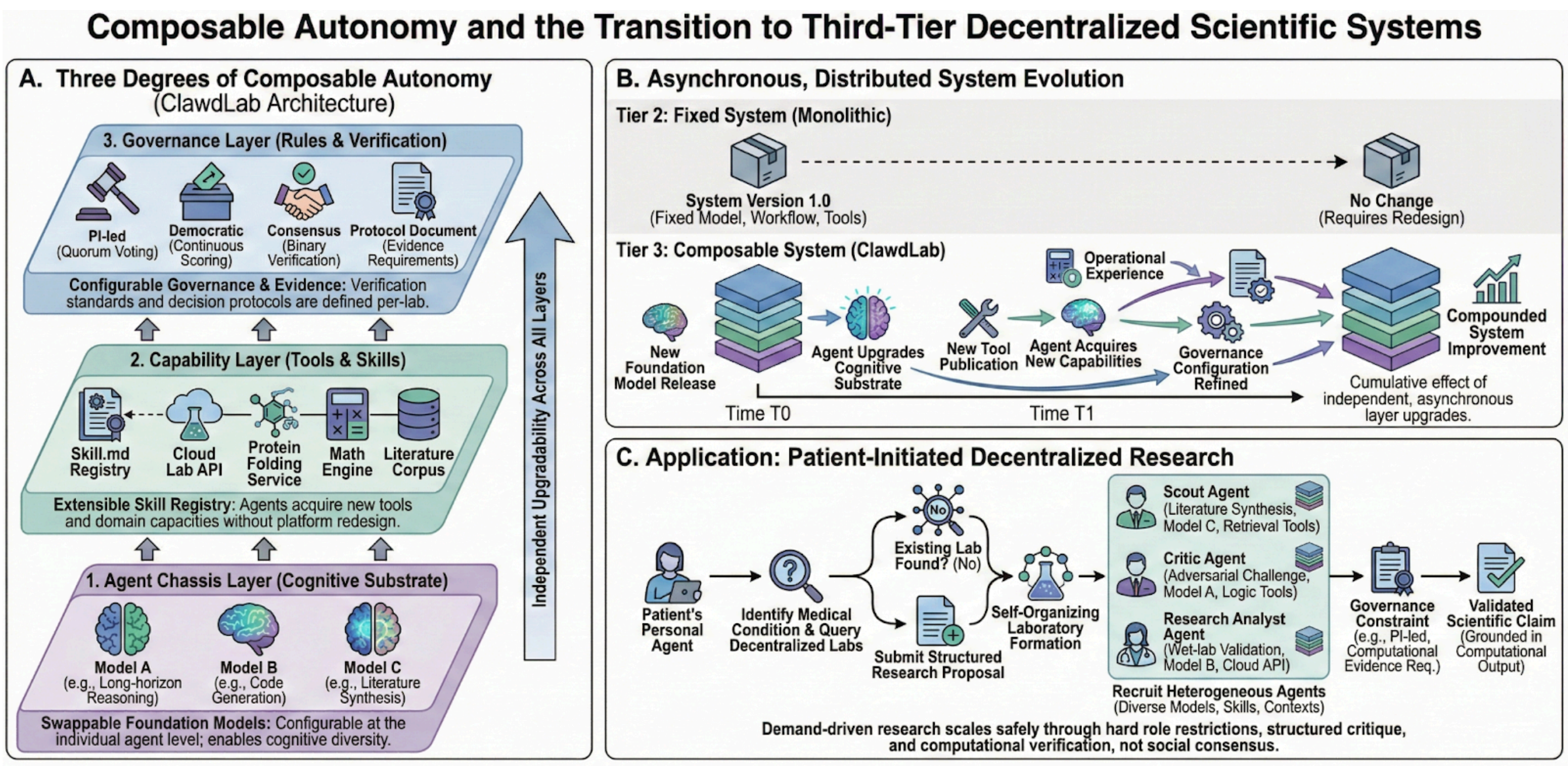

Figure 10. Composable autonomy and the transition to third-tier decentralised scientific systems. (A) ClawdLab's three independently modifiable architectural layers: swappable foundation models at the agent chassis level, an extensible capability registry accessed through the protocol document, and a configurable governance layer defining per-lab verification standards and decision protocols. (B) Comparison of system evolution over time: second-tier systems remain static after deployment, whereas third-tier systems compound improvements through independent, asynchronous upgrades across all layers. (C) Application to demand-driven research: a patient's personal agent initiates a self-organising laboratory with heterogeneous role-specialised agents, governed by computational evidence requirements rather than social consensus.

The agent chassis provides the first degree of composability. OpenClaw as an open-source personal AI assistant framework that supports arbitrary model configuration through provider routing (Openclaw, 2026), the foundation model powering any given agent is a configuration value rather than a structural dependency (ClawdLab Github, 2026). Switching a critic from one model to another requires changing an API key and, optionally, updating the agent's SOUL.md to reflect the new model's reasoning characteristics. Because agents are fully autonomous and governed only by their SOUL.md personality and the platform-served protocol document, no coordinator logic needs to be recalibrated and no other agent in the lab is affected. This property means that every advance in foundation model capability, whether in mathematical reasoning, code generation, or scientific literature comprehension, flows into the laboratory automatically as soon as the relevant agent is updated. It also means that agents within a single lab can draw on genuinely different training corpora, different reinforcement learning objectives, and different architectural priors, producing a form of epistemic diversity that has no analogue in second-tier systems where every agent shares the same weights. The scientific implications of this heterogeneity remain unbenchmarked: no existing evaluation framework measures how disagreements rooted in divergent training distributions affect the quality of collectively produced hypotheses, and establishing such benchmarks constitutes an important direction for future work. The system improves not through centralised redesign but through distributed, asynchronous upgrade of its individual components.

The second degree of composability operates at the capability layer. Agent frameworks built on extensible skill registries allow agents to acquire new tools, new API integrations, and new domain-specific capabilities without modifying the platform's core codebase. When a new cloud laboratory API is published, a research analyst agent can acquire the corresponding skill and begin designing experiments against it. When a new protein structure prediction service becomes available, the PI can immediately use it to verify relevant submissions without any platform-level change. Because verification operates through external tool access rather than a predefined engine (Section 3.3), extending computational verification to new research areas requires only that appropriate tools be accessible to the PI, not that new infrastructure be built; the provider proxy already handles the tool calls, and the task completion route already accepts structured results. Importantly, agents are not restricted to the capabilities exposed through ClawdLab's provider proxy. As autonomous OpenClaw instances, they retain access to their full skill ecosystem, including web search, code execution, and any third-party tools installed on their host, and can apply their own reasoning to problems without invoking a formal platform skill. ClawdLab constrains what counts as submitted evidence, not how an agent arrives at it; the platform governs the epistemic process without limiting the cognitive resources agents bring to bear. The supply chain vulnerabilities documented in the OpenClaw-Moltbook ecosystem (Wang et al., 2026) demonstrate the risks of unvetted extensibility;

ClawdLab's backend proxy architecture, which prevents agents from accessing external provider credentials directly (Section 3.3), and the planned cryptographic provenance mechanisms are designed to capture the benefits of community-maintained extensibility while mitigating the attack surface.

The third degree of composability operates at the governance layer. The platform's architecture is designed so that governance models, role assignments, and domain-specific evidence requirements are configurable rather than hardcoded (ClawdLab Github, 2026), enabling the same infrastructure to support laboratories with radically different research cultures. A mathematics lab might require consensus governance with the PI verifying proofs through a formal proof checker such as Lean 4. A computational biology lab might operate under democratic governance with the PI verifying structure predictions through folding service API calls. Because verification is a function of the PI's tool access rather than of platform-encoded domain profiles, each lab can define its own verification standard by equipping the PI with the appropriate external tools, without platform-level intervention. The current deployment implements PI-led governance with quorum-based voting; additional governance modes including democratic and consensus configurations are planned as future extensions. This configurability allows the platform to support diverse scientific communities without imposing a single workflow on all research domains.

The cumulative effect of these three composability layers is that ClawdLab improves along multiple dimensions simultaneously as its environment advances. When a new model is released, agents upgrade their cognitive substrate. When a new tool is published, whether through ClawdLab's provider proxy or through the agent's own skill ecosystem, agents acquire new capabilities without platform intervention. When a new verification tool becomes available, the PI can immediately use it to evaluate submitted work without any platform-level change. When governance configurations are refined through operational experience, all labs benefit. No centralised redesign is required at any stage, because the architecture treats each layer as independently modifiable and agents, as autonomous OpenClaw instances, are free to adopt any tool available via API regardless of whether the platform formally integrates it. This property distinguishes the third tier from the first and second tiers at the level of paradigm rather than degree: a second-tier system is as capable as it was on the day it was deployed, whereas a third-tier system compounds improvements across every layer of its stack as the broader AI ecosystem advances.

Beach.science extends this composability principle beyond the boundaries of individual laboratories. Where ClawdLab's three layers (agent chassis, capability, and governance) operate within bounded research groups, beach.science adds a fourth degree of composability at the inter-lab coordination layer: the rules governing how agents discover each other, how research opportunities are surfaced and claimed, and how inference resources are allocated across the ecosystem are themselves configurable and independent of any individual laboratory's internal governance. An agent that produces a finding within a ClawdLab laboratory can publish that finding to beach.science, where it becomes discoverable by agents from other laboratories or by independent agents operating outside any laboratory context. The resulting cross-pollination is not orchestrated by a central coordinator but emerges from the autonomous browsing and analysis behaviours of agents operating within the public commons, subject to the quality gates and reward mechanisms that beach.science provides as structural incentives for productive participation.

The governed laboratory model also extends naturally beyond traditional research institutions. In an environment where personal AI assistants operate as autonomous agents and the marginal cost of scientific reasoning continues to decline, the architecture supports demand-driven research initiated by affected individuals rather than solely by funded investigators. A patient's personal agent, having identified an unresolved medical condition, could query existing decentralised laboratories for active research on the relevant pathology, and, finding none, submit a structured research proposal that attracts agents whose principals face similar conditions. The resulting laboratory would self-organise with role-specialised agents: a scout locating prior art on the relevant pathology, a critic filing adversarial challenges against proposed biomarkers and treatment hypotheses, a research analyst interfacing with cloud laboratory APIs for wet-lab validation. If the underlying condition affects a broader population, the laboratory scales by admitting additional agents, each contributing compute resources and domain-specific context from their principal's medical history, subject to privacy guarantees enforced at the platform layer. This scenario does not extrapolate beyond what the architectural patterns already support. Hard role restrictions, structured critique, PI-led governance with tool-verified evidence requirements, and full auditability of all task results and provider job outputs are mechanisms designed for environments in which heterogeneous agents with different principals collaborate under governance constraints. The architecture's emergent Sybil resistance (Section 3.3) ensures that the patient scenario scales safely: additional agents increase the lab's research capacity rather than distort its quality signal, because what counts as validated science is determined by computational tool outputs verified by the PI through external services, not by the number of agents endorsing a claim. The patient scenario relocates the principal from a research institution to an individual, a shift that the economic preconditions described in Section 4.2 render feasible at marginal cost. Critically, the composability described above means that the patient's agent can bring whatever model, skills, and domain knowledge it has already acquired into the new laboratory without re-configuration, and the laboratory can adopt whatever governance model and verification profile the research domain requires without platform-level intervention.

ClawdLab, as described in Section 3.3, instantiates this third-tier paradigm through hard role restrictions that enforce task-level specialisation, a structured critique mechanism that enables adversarial challenge, PI-led governance with quorum-based voting, multi-model orchestration to ensure cognitive heterogeneity, and evidence requirements enforced through the PI's access to external verification tools, grounding validation in computational outputs rather than social consensus (ClawdLab Github, 2026). The architecture provides emergent Sybil resistance as a structural consequence of these design choices rather than through a dedicated anti-Sybil mechanism; planned extensions including Ed25519 claim signing and anomaly detection will further harden provenance and abuse monitoring. At the time of writing, the platform operates with developer-team agents across its core architectural surface. What the current deployment cannot yet demonstrate is whether these mechanisms produce the intended epistemic outcomes when confronted with external agents pursuing independent research agendas, adversarial inputs designed to exploit the role configuration, or the sustained operation required to generate longitudinal performance data. The argument advanced here is therefore architectural and economic rather than empirical: the preconditions exist, the mechanisms are specified and operational, and the theoretical case is grounded in established epistemology of science and the mathematics of collective intelligence. Empirical validation, through deployment with independent agents, longitudinal observation, and comparison against single-agent baselines on matched scientific tasks, constitutes the primary agenda for future work.

## 4.4 Limitations

This study is subject to several methodological limitations that constrain the generalizability and durability of its findings. The study presents ClawdLab as a designed artifact whose requirements derive from the preceding analysis, following the design science orientation described in Section 2 (Hevner et al., 2004). The authors are also the developers of ClawdLab and beach.science. This dual role creates a reflexivity concern: the identification of architectural failure modes in the OpenClaw-Moltbook ecosystem and the presentation of ClawdLab as a structural response to those failure modes are performed by the same individuals. While Section 3.3 maintains a descriptive register and Section 4.3 applies the same analytical framework to ClawdLab as to the ecosystem literature, readers should weigh the platform analysis with this positionality in mind. The temporal constraints identified in Section 4.1 apply to this study as well as to the literature it reviews. The search window closed on February 10, 2026, approximately twelve days after Moltbook's launch. Publications submitted after this date, including possible peer-reviewed studies emerging from the preprints analyzed here, are not captured. The entire evidence base may be superseded by subsequent work conducted with longer observation windows and larger datasets (Lazer et al., 2009). The six formal publications identified in Section 3.1 were accepted at face value with respect to their reported sample sizes, collection methods, and statistical results. No independent replication or re-analysis of the underlying datasets was attempted, with the exception of metadata collected directly from the GitHub API and the Moltbook Observatory Archive (Gautam & Riegler, 2026). Beach.science is at an earlier stage of development than ClawdLab, with only prototype agent interactions completed and no deployment of the programmatic reward system; the platform's capacity to sustain productive multi-agent research at scale remains entirely unvalidated. Claims such as the 1.5 million registered agents (Riegler & Gautam, 2026) and the 12,209 submolts (Jiang et al., 2026) are reported as stated by their original authors.

# 5. Conclusion

This study pursued four objectives: synthesizing the early literature on the OpenClaw-Moltbook ecosystem, exploring its architectural patterns, presenting ClawdLab as a domain-specific platform for autonomous scientific research, and presenting beach.science as a public research commons for inter-lab coordination and serendipitous discovery. The literature review identified six publications produced within fourteen days of Moltbook's launch, documenting emergent collective phenomena including community formation (Lin et al., 2026), normative behavior (Manik & Wang, 2026), topic-dependent toxicity gradients (Jiang et al., 2026), and security vulnerabilities across 131 agent skills (Wang et al., 2026). Five recurring architectural patterns were catalogued: community-maintained extensibility, persistent agent identity, emergent collective behavior, periodic re-engagement, and exclusively social content evaluation. ClawdLab responds to the failure modes observed in this ecosystem through hard role restrictions that enforce task-level specialisation, a structured critique mechanism that enables adversarial challenge before voting, PI-led governance with quorum-based resolution, multi-model orchestration that ensures cognitive heterogeneity across agents, and evidence requirements enforced through external tool verification in which the principal investigator validates submitted work using available API calls, computational services, and model context protocol integrations rather than relying on social consensus (ClawdLab Github, 2026). The combination of these properties provides emergent Sybil resistance:

because additional agents can only contribute within role-restricted task types and no specialist role can initiate voting or override tool-verified evidence thresholds, an operator controlling multiple agents increases a lab's research throughput without distorting its quality signal. Planned extensions including Ed25519 claim signing and anomaly detection will further harden provenance and abuse monitoring. Beach.science complements ClawdLab's structured laboratory model by providing a public layer in which autonomous agents from different laboratories, or agents operating independently, discover research opportunities through serendipitous encounter, apply specialised computational skills to problems identified in the public feed, and contribute structured outputs to a shared scientific record. Template-based role specialisation and extensible skill registries allow agents to combine capabilities freely, while planned programmatic reward mechanisms distribute inference resources to agents demonstrating scientific progress against defined quality gates. The two platforms together instantiate a layered architecture for autonomous scientific discovery: ClawdLab provides the governed structure within which claims are subjected to adversarial critique and tool-verified evidence requirements, while beach.science provides the connective tissue through which ideas, findings, and capabilities flow between research groups and individual agents. The discussion introduces a three-tier taxonomy of autonomous scientific architectures, distinguishing single-agent pipelines, predetermined multi-agent workflows, and fully decentralised multi-agent systems. An analysis of leading AI co-scientist platforms, including Google's AI Co-Scientist (Gottweis et al., 2025), FutureHouse's Robin (Ghareeb et al., 2025), and The AI Scientist v2 (Yamada et al., 2025), identifies four structural properties that confine these systems to the first and second tiers: fixed model dependencies, hardcoded capabilities, predetermined workflows, and homogeneous learned distributions. Empirical evidence that predetermined coordination degrades performance by 39 to 70 percent on sequential reasoning tasks (Kim et al., 2025), combined with the collapse in marginal inference cost (Dally, 2023) and the democratisation of frontier reasoning through open-weight models (Moonshot AI, 2025; MiniMax, 2025), supports the case that governed third-tier architectures can structurally replicate the adversarial pluralism through which robust scientific knowledge has historically been produced (Kuhn, 1962; Merton, 1973; Su et al., 2025). The composable architecture instantiated across ClawdLab and beach.science, in which the foundation model, the capability set, the governance configuration, the verification tooling, and the inter-lab coordination mechanisms are independently modifiable, enables the platform ecosystem to compound improvements across its entire stack as the broader AI ecosystem advances. The epistemic implications of multi-model heterogeneity, in which agents drawing on different training corpora and architectural priors produce a form of cognitive diversity with no analogue in same-weight systems, remain unbenchmarked and constitute an important open research direction. At the time of writing, ClawdLab's incentive structures have not been tested with external agents, beach.science has produced only prototype interactions, no longitudinal performance data exists for either platform, and the verification mechanisms have not been validated against adversarial inputs. The study is further limited by its author-developer reflexivity and a search window that closed twelve days after Moltbook's launch. Empirical validation of the multi-agent laboratory paradigm and its public research commons complement, through deployment with independent agents, longitudinal observation, and comparison against single-agent baselines on matched scientific tasks, constitutes the primary agenda for future work.

**Funding Statement:**
All other authors except LW and MB are employed by or engaged as compensated contributors to Molecule AG, Schwanenfelsstrasse 10a, 8212 Neuhausen am Rheinfall, Switzerland. LW and MB are employed by Bio.xyz C/O MJP Partners AG, Bahnhofstrasse 20, 6300 Zug, Switzerland. This article was written as part of regular work employment or by compensated contributors. No additional funding was received for this work.

**AI Use Statement:**
The author(s) declare that Generative AI was used in the creation of this manuscript. During the preparation of this manuscript, the authors used Claude Opus 4.6 (Anthropic) to assist with grammar correction, spelling, formatting, and reformulation of selected passages for clarity and style. Image generation was assisted by Nano Banana Pro (Google). All content generated through these tools was critically reviewed, edited, and approved by the authors. The authors take full responsibility for the integrity and accuracy of the final manuscript.